\documentclass{article}

\usepackage{arxiv}

\usepackage[utf8]{inputenc} 
\usepackage[T1]{fontenc}    
\usepackage{hyperref}       
\usepackage{url}            
\usepackage{booktabs}       
\usepackage{amsfonts}       
\usepackage{amsmath}        
\usepackage{nicefrac}       
\usepackage{microtype}      
\usepackage{lipsum}		
\usepackage{graphicx}
\usepackage{epstopdf}
\usepackage{natbib}
\usepackage{doi}

\title{The Infinite-Dimensional Nature of Spectroscopy and Why Models
Succeed, Fail, and Mislead}

\author{ \href{https://orcid.org/0000-0002-6060-5365}{\includegraphics[scale=0.06]{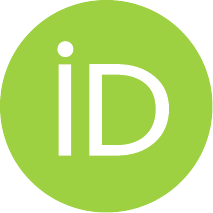}\hspace{1mm}Umberto Michelucci} \\
	Department of Computer Science\\
	Lucerne University of Applied Sciences and Arts\\
	Risch-Rotkreuz, 6343 \\
    Switzerland \\
	\texttt{umberto.michelucci@hslu.ch} \\
	\And
	\href{https://orcid.org/0000-0003-2562-9932}{\includegraphics[scale=0.06]{orcid.pdf}\hspace{1mm}Francesca Venturini} \\
	Institute of Applied Mathematics and Physics\\
	ZHAW Zurich University of Applied Sciences\\
	Winterthur, 8400 \\
    Switzerland \\
	\texttt{vent@zhaw.ch} \\
}

\renewcommand{\headeright}{}
\renewcommand{\undertitle}{Published in The Analyst, Royal Society of Chemistry}
\renewcommand{\shorttitle}{The Infinite-Dimensional Nature of Spectroscopy and Why Models Succeed, Fail, and Mislead}

\hypersetup{
pdftitle={The Infinite-Dimensional Nature of Spectroscopy and Why Models
Succeed, Fail, and Mislead},
pdfsubject={},
pdfauthor={Umberto Michelucci, Francesca Venturini},
pdfkeywords={Machine Learning, Infinite Dimensions, Spectroscopy, Classification, Feldman-Hajek Theorem},
}

\begin{document}
\maketitle

\begin{abstract}
	Machine learning (ML) models have achieved strikingly high accuracies in spectroscopic classification tasks, often without a clear proof that those models used chemically meaningful features. Existing studies have linked these results to data preprocessing choices, noise sensitivity, and model complexity, but no unifying explanation is available so far. In this work, we show that these phenomena arise naturally from the intrinsic high dimensionality of spectral data. Using a theoretical analysis grounded in the Feldman–H\'ajek theorem and the concentration of measure, we show that even infinitesimal distributional differences, caused by noise, normalisation, or instrumental artefacts, may become perfectly separable in high-dimensional spaces. Through a series of specific experiments on synthetic and real fluorescence spectra, we illustrate how models can achieve near-perfect accuracy even when chemical distinctions are absent, and why feature-importance maps may highlight spectrally irrelevant regions. We provide a rigorous theoretical framework, confirm the effect experimentally, and conclude with practical recommendations for building and interpreting ML models in spectroscopy.
\end{abstract}

\keywords{Machine Learning, Infinite Dimensions, Spectroscopy, Classification, Feldman-Hajek Theorem}

\section{Introduction}\label{sec1}

Spectroscopy is the study of how matter interacts with electromagnetic radiation, typically by measuring the intensity of emitted light as a function of wavelength or frequency. The analysis of these spectral signatures non-destructively can reveal the composition and then chemical structure of a sample.

Extracting chemical and physical information from spectra is usually complex and requires involuted data processing pipelines that include steps such as, for example, baseline subtraction or smoothing \cite{mark2010chemometrics,morais_tutorial_2020,guo2021chemometric}. 
Interpreting the output of machine learning (ML) models applied to spectra presents several challenges: the high number of wavelengths complicates interpretation, complex models might capture nonlinear interactions, making it difficult to connect features in spectra with chemical information about the sample (models are often black-boxes, for lack of interpretability). Complex models may fit noise rather than signal, making predictions useless \cite{noauthor_demystifying_2025}. Furthermore, it has become clear to the research community that attributing a prediction to specific wavelength bands does not have a unique solution, and different methods lead to different attributions \cite{mamalakis_carefully_2022}. This is due to the fact that explainability approaches measure how a specific model responds to changes in intensity at individual wavelengths, which very often include regions far from chemically significant peaks that the model learns to exploit due to subtle statistical differences \cite{contreras_explainable_2025}. Zehtabvar \textit{et al.}  \cite{zehtabvar_review_2024} have found that data normalisation has a strong influence on the accuracy of ML models, something that seems strange, since data normalisation does not have a relationship with physico-chemical information about measurements. Contreras \textit{et al.} \cite{contreras_explainable_2025}, clearly show how feature importance algorithms are susceptible to noise-induced fluctuations, although they fail to give a good explanation for this observation.  They also note that because of the high dimensionality of the data, interpretation of the results is challenging.

Steinmann \textit{et al.} \cite{steinmann_navigating_2024} studied the problem of spurious correlation in a long article and compared it to the ``clever Hans behaviour''. The term \textit{Clever Hans} comes from animal psychology, named after the horse Hans that apparently had learnt to understand human language. Hans (the horse) instead learnt to rely on the facial expressions of humans asking questions and was unable to give correct answers when not seeing the human face (for more interesting details on the case, you can read \cite{samhita_clever_2013}). To paraphrase Steinmann \textit{et al.}, in some cases, ML algorithms in spectroscopy are as dumb as a horse\footnote{The authors do not want to imply that horses are not intelligent animals, only that they cannot classify spectra accurately.}.
The fact that ML is seemingly capable of classifying any spectra dataset with a high accuracy has sparked the appearance of uncountable articles that use ML to extract the elusive chemical or physical parameters sought without the need of involuted data analysis pipelines (see for an overview \cite{houhou_trends_2021}).

How such models really generalise to new measurement setups or datasets is an open question that cannot be answered uniquely today. 
This depends on whether a model learns from physically meaning features (e.g., an absorption or emission line), or from artefacts of the measurement process (e.g., from the noise introduced by a specific electronics \cite{alkemade_review_1978}). In the first case, the model is likely to generalise to new measurements well, whereas in the latter case the model simply overfits specific characteristics of the measurement apparatus, rendering it unreliable.

This article, for the first time, explains why the high dimensionality of spectroscopy (the number of intensity values in spectra is usually of the order of 10$^3$) is responsible for the ability of ML to classify almost all kinds of spectroscopy dataset, even in situations where the data itself contain no discernible feature to distinguish between classes (e.g., all intensities in spectra in one class noticeably higher than in another class). Fundamentally, this work demonstrates that the effectiveness of ML models when applied to spectroscopy may be due in many cases to the high dimensionality of the spectral data rather than chemical-physical spectral features. Especially flexible models, such as random forests, can obtain an almost perfect classification accuracy even using spectral regions that do not contain relevant physico-chemical features because of the high dimensionality of the data. This work demonstrates that, in high-dimensional spaces, even subtle differences in the statistical properties of the signal across classes can enable a model to achieve seemingly perfect classification accuracy, despite the spectra themselves lacking sufficient information to justify such performance. 
This may be particularly relevant in reflectance and fluorescence spectroscopy, where the spectra usually have broad features rather than sharp specific signatures, such as in Raman-spectroscopy.


The contributions of this article are the following. (i) We present a mathematical discussion of the role of high dimensionality in finite- and infinite-dimensional cases, with a specific discussion of how to translate those abstract results to spectroscopy. (ii) We present a series of experiments on synthetic data to show under which conditions high-dimensionality becomes relevant in classification. (iii) We show how this phenomenon appears in real fluorescence spectroscopy data. 
(iv) Finally, we describe how spectroscopists should change their way of analysing spectra to take into account the effect of high dimensionality.


This article is structured in the following way. In Section \ref{sec:methods} we discuss the mathematical reasons for the behaviour of Gaussian distributed data in finite- and infinite-dimensional spaces (the Feldman-H\'ajek theorem \cite{feldman1958, hajek__1958}). We then proceed to generalise the findings for the non-Gaussian distributed data and explain why it is applicable to spectroscopy data. In Section \ref{sec:exp} we describe a series of experiments on synthetic data and real data used to investigate the effect of dimensionality. In Section \ref{sec:results} we present the results obtained on synthetic and real data. In Section \ref{sec:spectroscopy}, we discuss the relevance of the results for spectroscopy. Finally, in Section \ref{sec:concl} we discuss conclusions and limitations. 

\section{Methods}
\label{sec:methods}

To study how the classification accuracy of models varies when applied to spectral data as a function of the number of intensities (referred to in this paper as the spectrum" dimensions), it is useful to begin with the most basic case: data that follow a normal distribution. Later in this paper, we will expand the discussion to more realistic cases that reflect spectroscopy data more closely.

\subsection{Gaussian Case}

The key mathematical theorem that describes under which conditions the data distributed according to two different Gaussians are separable (in other words, perfectly classifiable) is the Feldman-H\'ajek thorem, proved in 1958 by Feldman \cite{feldman1958} and H\'ajek independently\cite{hajek__1958}.
This theorem is well known in Gaussian measure theory, but less in applied machine learning circles. 
The theorem provides the conditions under which two Gaussian measures are either \textit{mutually absolutely continuous} or \textit{mutually singular} (to better understand the meaning of these two concepts, the reader is referred to \cite{axler_measure_2020}). In intuitive terms, it tells when two sets of data, $X_1$ and $X_2$, drawn from \textit{different} multivariate normal distributions, can be regarded as perfectly separable from a measure-theory point of view (which means that there exists a possibly very complex classifier that can separate them with zero classification error).

The implications of the Feldman--H\'ajek theorem are highly relevant for spectroscopy. A spectrum can be viewed as a point in a high--dimensional space, where intensity at each wavelength (or pixel) represents one coordinate. When this dimensionality is large (for example, 10$^3$ intensities values), the \textit{geometry} of the space in which spectra are defined changes dramatically. 
The theorem states that, under assumptions verified in the case of spectroscopy, in finite dimensions, two Gaussian distributions with slightly different means or variances always overlap to some extent and can never be perfectly classified. In contrast, in infinite (or with a good approximation in very high) dimensions, even the smallest difference in mean or covariance makes the two distributions \emph{mutually singular}, meaning that they occupy disjoint regions of the space, and as such, they can be perfectly classified by an appropriate algorithm. 
For a spectroscopist, this provides a rigorous explanation for a common observation:
ML models often achieve a very high accuracy even when spectra appear indistinguishable. The Feldman--H\'ajek theorem shows that this behaviour is a
\textit{geometric} consequence of high dimensionality: tiny instrumental artefacts, baseline shifts, or preprocessing differences can make two classes of spectra perfectly
separable, even in the absence of any genuine physico-chemical distinction. Thus, the theorem clarifies why models may ``succeed'' mathematically while not learning from physico-chemical meaningful information.

\subsection{Non Gaussian Case}
\label{sec:non-gaussian}

Spectral data typically do not follow a normal (Gaussian) distribution. Consequently, one might initially assume that the previous discussion does not adequately describe the behaviour of real-world datasets. The Feldman–Hájek theorem can be generalized\cite{michelucci2026feldmanhajek} to the case of an infinite countable collection of Gaussian mixture models (in what is called the Gaussian mixture dichotomy theorem). Thus, since the distribution of any data can be approximated to an arbitrarily high degree of accuracy by such mixtures, this extension effectively applies to essentially any dataset. That substantially means that in the limit of infinite dimensions we can expect an almost perfect classification for basically every dataset (up to certain limits that depend on the chosen classifier and dataset).


\subsection{Concentration of Measure}

A further justification of the result comes from the phenomenon known as \emph{concentration of measure}. To intuitively understand it, consider the following example.
In our everyday 3-dimensional world, a solid orange is mostly ``fruit'' and very little ``peel''. If you were to thin the peel slightly, the amount of fruit inside would not change much. However, in high-dimensional spaces (such as the 1,024-pixel space of a spectrometer), geometry works backward. An orange in 1000 dimensions (or more) will be almost entirely concentrated on the peel, and it will be almost completely empty inside (that would make peeling an orange quite an endevour, so we should probably be thankful not to live in 1000 dimensions).

Basically, as you add dimensions, the "volume" of a shape migrates away from the centre and traps itself almost entirely in the outer shell. In 1,024 dimensions, a "solid" ball (your organge) is essentially empty; 99.9\% of its contents exist only in a paper-thin layer on the surface.
This means that if you pick a random point that is part of the ball, its distance from the centre will almost always be the same, since it will be with almost certainty on the shell (the orange peel) (this is something that defies completely intuition)!  
In other words, nearly all points of the ball have a norm $||x||_2$ (the length of the vector from the origin to a point $x$ part of the ball) close to a typical value.

\begin{figure}[hbt]
    \centering
    \includegraphics[width=0.75\linewidth]{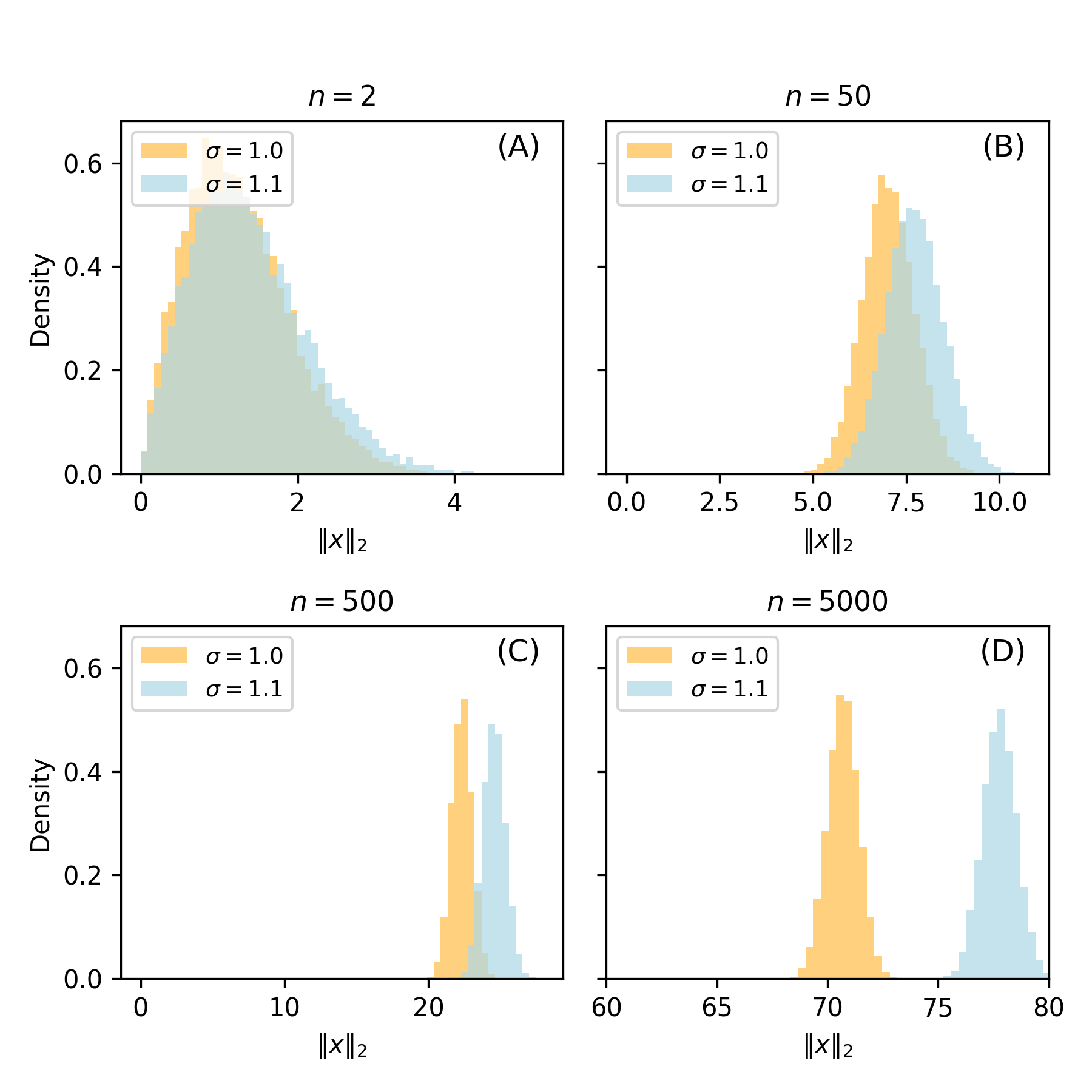}
    \caption{Illustration of the concentration of measure for multivariate Gaussian
distributions. Shown are the empirical distributions of $\|x\|_2$ for samples drawn
from $\mathcal{N}(0,\,1.0^2 I_n)$ (light blue) and $\mathcal{N}(0,\,1.1^2 I_n)$
(yellow), for increasing dimensionalities $n=2$, $50$, $500$, and $5000$
(panels~A–D). In low dimensions the two distributions overlap substantially, but as
$n$ increases the probability mass concentrates sharply around the typical radius
$\sigma\sqrt{n}$, and even small variance differences cause almost complete
separation. This illustrates how, in high-dimensional spaces, measures supported by
Gaussian (and many non-Gaussian) distributions become effectively disjoint, providing
an intuitive geometric basis for the Feldman--H\'ajek theorem and for the sensitivity
of high-dimensional classifiers to minute statistical differences.}
    \label{fig:concentration}
\end{figure}

This phenomenon shows itself in high-dimensional spaces, and the probability mass (intuitively the values of the norm of the arrays $||x||_2=(x_1^2+\cdots+x_n^2)^{1/2}$) for a Gaussian distributed dataset in $n$ dimensions tends to concentrate around $\sigma \sqrt{n}$ (assuming for simplicity that $x\sim \mathcal{N}(0, \sigma^2 I_n)$). 
For a visualisation of this phenomenon, in Figure \ref{fig:concentration} we show the distributions of $||x||_2$ (the length of vectors $x$) sampled from two Gaussian distributions $\mathcal{N}(0, 1.0^2I_n)$ and $\mathcal{N}(0, 1.1^2I_n)$ where $I_n$ is the identity matrix of dimensions $n\times n$. For illustrative reasons, we consider the case of isotropic covariances, but this phenomenon is still happening for generic covariances.
In panel (A) of Figure \ref{fig:concentration} it can be seen that in dimension 2, the $||x||_2$ distributions have a high overlap (as intuitively clear since the two distributions have the same mean and only slightly different covariances).
As the dimension increases (panel (B), (C), and (D)), distributions overlap decrease, until the dimensionality is high enough (panel (D)), and the two have almost no overlap anymore.

In general, random variables with finite variance exhibit extremely small relative fluctuations even when not Gaussian: most realisations (the measured values) lie very close to their expected value. 
This implies that the geometry of high-dimensional data is effectively governed by its first- and second-order statistics (mean and covariance) and that differences in these quantities dominate the behaviour of distances and overlaps between distributions. 
Hence, the Gaussian assumption underlying the Feldman--H\'ajek theorem remains a valid guide for understanding separability and equivalence of high-dimensional or averaged data, 
even when the underlying distributions deviate from strict normality.

\subsection{Effect in Ohter Fields}

The ``dimensional traps'' identified in this work are not unique to spectroscopy; they represent a fundamental challenge across high-dimensional measurement sciences. In genomics, for instance, the $p \gg n$ problem (where the number of genes far exceeds the number of patients) frequently leads to `phantom' biomarkers \cite{ioannidis2005microarrays, clarke2008properties} that fail to replicate in clinical trials—a phenomenon often attributed to the model's ability to find a separating hyperplane in the high-dimensional noise floor of the microarray or sequencing data. Similarly, in functional MRI (fMRI) neuroimaging, researchers have documented 'voodoo correlations' \cite{vul2009voodoo} (to use the same words that Vul \textit{et al.} used) where high-dimensional voxel-wise patterns yield near-perfect classification of psychological states, only to be later identified as artefacts of head motion or instrumental sampling bias.

\section{Experiments}
\label{sec:exp}

To show convincingly that high-dimensionality may be responsible for the exceptional performance of ML models in spectroscopy, a series of experiments on synthetic datasets and on a real fluorescence dataset are presented in this article. 

The overview of the experiments is presented in Table \ref{tab:exp-overview}. 
Experiments are indicated with N1, N2, N3, and N4 for noise classification experiments, with S1, S2, and S3 for synthetic spectra classification experiments, and with R1a to R5b for real data classification experiments.

\begin{table}[hbt]
\caption{Overview of experiments and their core ideas.}
\label{tab:exp-overview}
\centering\footnotesize
\setlength{\tabcolsep}{4pt}
\begin{tabular}{@{}p{.05\textwidth}p{.3\textwidth}p{.6\textwidth}@{}}
\toprule
\textbf{ID} & \textbf{Experiment} & \textbf{Experiment Details} \\
\midrule \midrule

N1 & Gaussian noise: $\Delta\sigma$ sweep (QDA)
& Classify white noise from two multivariate Gaussians with equal means ($\mu_1=\mu_2=1$) and different isotropic and Toeplitz (with parameter $\rho=0.95$) covariances; we varied the variance gap $\Delta\sigma:=|\sigma_2-\sigma_1|$ from 0 to 2 while keeping $\sigma_1=1$ fixed. We have tested $n=\{ 5,10,50,500\}$. We have measured classification accuracy. \\

\midrule
N2 & Gaussian noise: Bayes “oracle” boundary & Classify white noise from two multivariate Gaussians with equal means ($\mu_1=\mu_2=1$) and different isotropic covariances; we varied the variance gap $\Delta\sigma:=|\sigma_2-\sigma_1|$ from 0 to 2 while keeping $\sigma_1=1$ fixed. We have tested $n=\{ 30, 100, 500, 1000, 5000\}$. 
We classify white noise by using the ideal threshold on $\sum (x_j-\mu)^2$. \\

\midrule
N3 & Gaussian noise: accuracy vs.\ dimension $n$
& Classify white noise from two multivariate Gaussians with equal means ($\mu_1=\mu_2=0$) and different isotropic covariances; we varied the variance gap in $\Delta\sigma:=|\sigma_2-\sigma_1|=\{ 0.1, 0.3, 0.6, 0.9, 1.2, 1.5, 2.0\}$ while keeping $\sigma_1=1$ fixed. We tested $n=\{ 1,5,10,20,50,100,200,500,1000,2000,5000\}$.  We calculate the accuracy of QDA for various values of $n$. \\

\midrule
N4 & Skew–normal noise: 2D parameter sweeps
& Classify two classes of white noise from multivariate skew–normal \cite{azzalini_multivariate_1996}. For this experiment, we used the parameters $n=50$, $\mu_1=\mu_2=10$, $\sigma_1=1$, $\gamma_1=0.5$. We then varied $\Delta \sigma/\sigma$ from 0 to 2, $\Delta \mu/\mu$ from 0 to 0.15, and $\Delta \gamma_1/\gamma_1$ from 0 to 8. \\

\midrule
S1 & Synthetic spectra: truly identical classes
& Classify two classes of spectra each composed of one Lorentzian (both centres chosen according to a normal distribution $\mathcal{N}(50, 10^2)$) and an FWHM $\xi=7$. We generated $N=500$ spectra for each class. We choose $n=\{ 5, 10, 50, 100, 1000, 2000, 5000, 10000\}$. These two sets of spectra are not distinguishable, as there are no differences in the data distributions.\\

\midrule
S2 & Synthetic spectra: FWHM difference
& Classify two classes of spectra each composed of one Lorentzian (both centres chosen accordingly to a normal distribution $\mathcal{N}(50, 10^2)$) and two different FWHM $\xi_1=7$ and $\xi_2=9$. We generated $N=500$ spectra for each class. 
We choose $n=\{ 5, 10, 50, 100, 1000, 2000, 5000, 10000\}$ We study accuracy of classification for various value of $n$. \\

\midrule
S3 & Synthetic spectra with additive noise offset
&Classify two classes of spectra each composed of one Lorentzian (both centres chosen accordingly to a normal distribution $\mathcal{N}(50, 10^2)$). We have chosen $n=\{ 5, 10, 50, 100, 1000, 2000, 5000, 10000\}$ and a FWHM $\xi=7$. We generated $N=500$ spectra for each class. We then added i.i.d.\ Gaussian noise with a tiny class-specific mean offset ($0$ vs.\ $0.01$) and the same standard deviation of 0.01 (namely the noise was chosen from the two distributions $\mathcal{N}(0,0.01^2)$ and $\mathcal{N}(0.01,0.01^2)$). \\
\midrule
Ra1/Rb1 & Global pixel permutation 
& A single, consistent random shuffle is applied to all pixels across the entire dataset. This destroys physical contiguity (peaks and baselines) while preserving the global covariance structure. This tests if the model relies on spectroscopic shapes or high-dimensional statistical geometry. \\

\midrule

Ra2/Rb2 & Independent row permutation 
& Every spectrum is shuffled using a unique random seed, destroying both physical contiguity and inter-pixel covariance. This serves as a control to demonstrate that model success vanishes when the high-dimensional statistical structure is eliminated. \\

\midrule
Ra3/Rb3 & Pixel count sweep 
& Classify EVOO vs. LOO (Ra1) and EVOO vs. VOO (Rb1) using an increasingly high number of randomly chosen pixels $k \in [2, 35]$ from the first 50 pixels (Region $\rho_1$, noise only). For each $k$, 20 independent random subsets were tested to evaluate the climb in accuracy within chemically empty regions. \\

\midrule
Ra4/Rb4 & Feature importance: window sweep 
& Classify oils using non-overlapping moving windows of increasing widths $W \in \{20, 50, 200, 400\}$ across the entire detector. This experiment tests if near-perfect separability persists in regions lacking physical signals (0--400 px) compared to peak regions (600--800 px). \\

\midrule
Ra5/Rb5 & Feature importance: SHAP
& Generate mean absolute SHAP attribution maps for both experiments (A and B) across different window sizes. This experiment identifies whether the model's "important" features correlate with chemical peaks or are distributed across the high-dimensional noise floor. \\

\bottomrule
\end{tabular}
\end{table}

\subsection{Gaussian Noise Classification}
\label{sec:gauss1}

As a first set of experiments (N1, N2, and N3), we generated two classes of random noise arrays of dimension $n$, sampled from an isotropic multivariate distribution in the range $\mathcal{N}(\mu_1, \sigma_1^2I_n)$ for class 1 and $\mathcal{N}(\mu_2, \sigma_2^2I_n)$ for class 2. These experiments aim to testing the accuracy of classifiers in distinguishing the two classes for various values of the standard deviation $\sigma_1$ and $\sigma_2$ varying dimensionality $n$, with different approaches: QDA (experiments N1 and N3) and LDA decision boundary (N2).
In N1 and N2 we have $\mu_1=\mu_2=1$, while in N3 we have $\mu_1=\mu_0=0$.
Choosing the covariance matrix as $\sigma^2 I_n$ is equivalent to saying that each intensity in the spectra (at each wavelength) is completely independent of the others. This is clearly not true. Although the Feldman-H\'ajek theorem is valid for a generic covariance (under certain assumptions, discussed in the appendices), it is an interesting question to ask what the effect of correlation between intensities at different wavelengths on the effect of high-dimensionality is. To study this, we performed the same experiments also with a Toeplitz geometric covariance.
The latter is a matrix whose entries depend only on the absolute difference between their indices. This type of covariance is modelled with a parameter $\rho \in (-1,1)$. For an $n$-vector $x=(x_1,\dots,x_n)$ we write $\Sigma(\rho,\sigma^2)\in\mathbb{R}^{n\times n}$ with entries
\begin{equation}
\Sigma_{ij} \;=\; \operatorname{Cov}(x_i,x_j) \;=\; \sigma^2\,\rho^{\,|i-j|},
\qquad 1\le i,j\le n,
\end{equation}
which is Toeplitz, since each diagonal is constant. This type of covariance models situations where the correlation between intensities decreases as the difference between their wavelengths increases. Although this is an approximation in spectroscopy, where there might be large covariances between wavelength bands, it is an approximation that gives an indication on the effect of correlation and that is easily tractable mathematically and numerically.

The details of the experiments with the complete ranges of parameters tested are contained in Table \ref{tab:exp-overview}.

\subsection{Skewed Normal Noise Classification}
\label{sec:skew}

As we discussed in Section \ref{sec:non-gaussian}, even when data are not normally distributed (as spectra are not), the high-dimensionality effect is still very powerful. To show this, we performed experiments using the skewed normal distribution (N4 in Table \ref{tab:exp-overview}). This choice was guided by the analysis of noise present in the real data discussed in Section \ref{sec:oil}.

For this experiment, we generated two classes of random noise arrays of dimensions $n$, analogously to the method described in the previous section, from the skewed normal distributions (SND) defined by Azzalini and Dalla-Valle \cite{azzalini_multivariate_1996} , described below. 

Let us give the mathematical form of the SND used. In its univariate version the probability density function (PDF) of a SND is given by 
\begin{equation}
    f(x) = 2\phi(x)\Phi(\alpha x)
\end{equation}
with 
\begin{equation}
    \phi(x)=\frac{1}{2\pi}e^{-x^2/2}
\end{equation}
the standard normal PDF and $\Phi(x)$ the cdf of the univariate standard normal distribution. For this work, we used the multivariate version of this distribution \cite{azzalini_multivariate_1996, mondal_multivariate_2024}.

The multivariate skewed normal distribution (MSN) introduced by Azzalini and Dalla-Valle  \cite{azzalini_multivariate_1996} is defined in its general form as
follows. A random vector $X\in\mathbb{R}^{p}$  has an SND distribution with location parameter $\boldsymbol{\xi}\in\mathbb{R}^{p}$,
symmetric positive definite scale parameter $\boldsymbol{\Omega}\in\mathbb{R}^{p\times p}$, and
skewness parameter $\boldsymbol{\alpha}\in\mathbb{R}^{p}$, if its multivariate PDF is 
\begin{equation}
  f_{X}(x)
  = 2\,\phi_{p}\!\left(x;\,\boldsymbol{\mu},\boldsymbol{\Omega}\right)\,
    \Phi\!\left\{\boldsymbol{\gamma}^{\top}\boldsymbol{\omega}^{-1}\,(x-\boldsymbol{\mu})\right\},
  \ x\in\mathbb{R}^{p}
  \tag{2}
  \label{eq:asn-pdf}
\end{equation}
where $\phi_{p}(\,\cdot\,;\boldsymbol{\mu},\Sigma)$ is the multivariate PDF of a $p$-dimensional normal distribution with mean $\boldsymbol{\mu}\in\mathbb{R}^{p}$  and $\boldsymbol{\omega}=\operatorname{diag}(\boldsymbol{\Omega})^{1/2}$. 
For completeness, we can write
\begin{equation}
    \phi_p(x;\boldsymbol{\Omega})
= (2\pi)^{-p/2}\,|\boldsymbol{\Omega}|^{-1/2}
  \exp\!\Big(-\tfrac12 (x-\boldsymbol{\mu})^\top
  \boldsymbol{\Omega}^{-1}(x-\boldsymbol{\mu})\Big).
\end{equation}
Note that the matrix $\boldsymbol{\Omega}$ is a dispersion matrix and equals the covariance of
$X$ only for $\boldsymbol{\alpha} = 0$.

For our tests, we used $\boldsymbol{\mu} = \mu  1_p$ and $\boldsymbol{\gamma} =\boldsymbol{\gamma} 1_p$ and $\boldsymbol{\Omega} = \sigma^2 I_p$. We then evaluated the separability of the classes with synthetic experiments. 
For each model, we generated two classes of $N=100$ samples in $n=50$ dimensions with coordinates i.i.d.\ from skew-normal distributions: a fixed base class $(\mu=10,\sigma=1,\gamma_1=0.5)$ and a perturbed class obtained by shifting parameters by $\Delta\mu$, $\Delta\sigma$, and $\Delta\gamma_1$ over uniform grids. 
We performed three two-dimensional sweeps, $(\Delta\mu,\Delta\sigma)$, $(\Delta\mu,\Delta\gamma_1)$, and $(\Delta\sigma,\Delta\gamma_1)$, and, at each grid point, trained a specific model and measured out-of-sample accuracy via 5-fold cross-validation. 
The mean cross-validated accuracies define accuracy surfaces that quantify how separability changes with differences in mean, variance, and skewness. 
For reproducibility, the two classes were generated with independent random seeds, and we also report results on the relative axes $(\Delta\mu/\mu,\Delta\sigma/\sigma,\Delta\gamma_1/\gamma_1)$. 

\subsection{Synthetic Spectra Classification}

To highlight the importance of high-dimensionality, we performed experiments with simulated spectra (S1, S2, and S3 in Table \ref{tab:exp-overview}). 
In a first experiment S1 we want to show how if spectra are truly undistinguishable, then no matter the dimension or the model, classification will be no better than chance.

To show this, we simulated two classes of one–peak spectra on a discrete axis
$x=(x_1,\ldots,x_{n})$ (representing detector pixels, wavenumbers or wavelengths depending on the spectroscopy type). 
Each spectrum is a Lorentzian profile with a randomly jittered centre and a fixed full width at half maximum (FWHM), equal for the two classes. 
For class $k\in\{1,2\}$ we draw, independently for every spectrum $j$,
a peak centre $c_j \sim \mathcal N(\mu,\sigma^2)$ (with values $\mu=50$, $\sigma =10$ and $\text{FWHM}=\xi = 7$ for numerical simulations). We used the unit-height Lorentzian.
\[
L(x;c,\xi)\;=\;\frac{(\xi/2)^2}{(x-c)^2+(\xi/2)^2},
\]
We generate $N=1000$ spectra per class with $n=100$. 

\begin{figure}[hbt]
    \centering
    \includegraphics[width=0.75\linewidth]{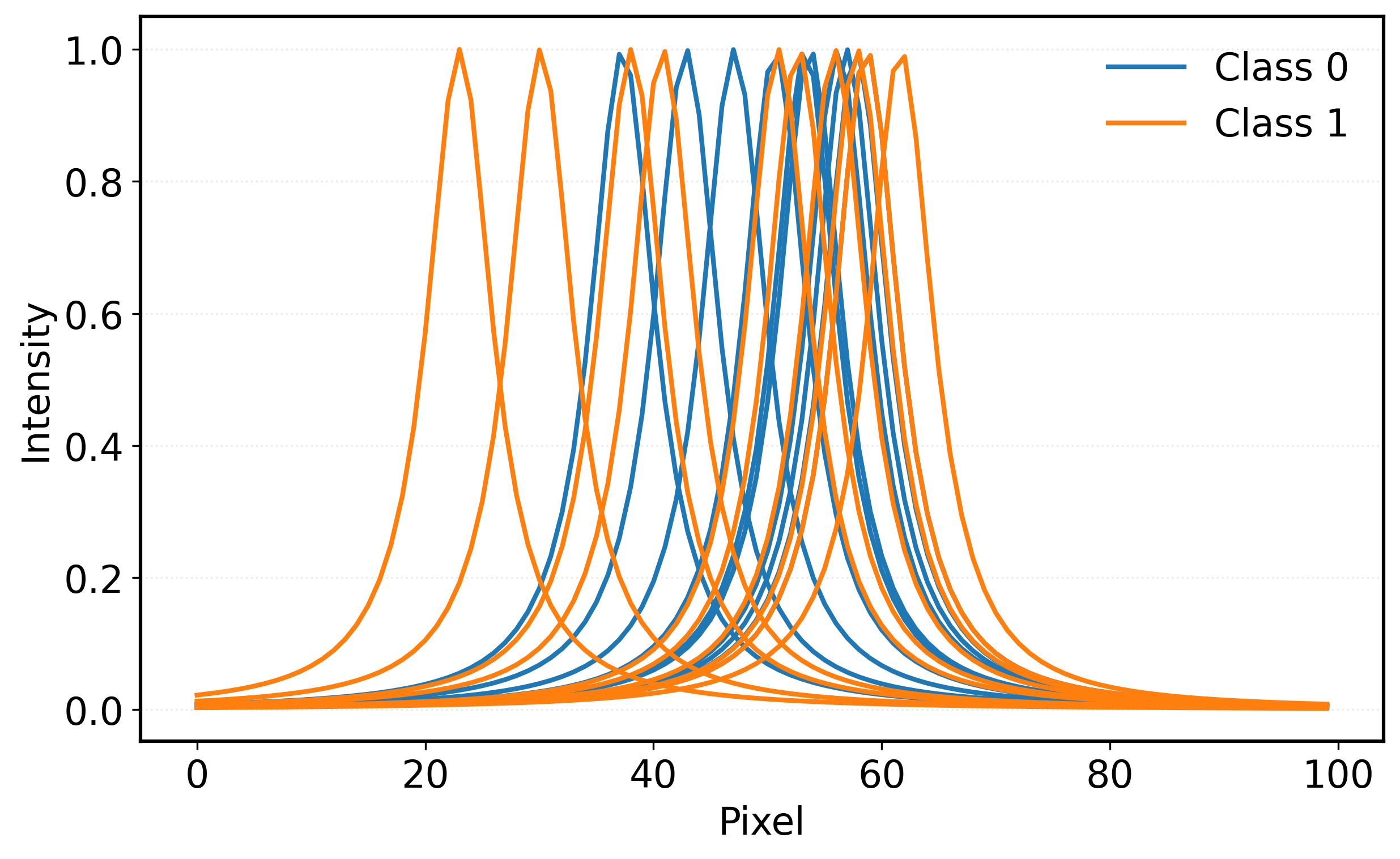}
    \caption{Ten representative synthetic one-peak spectra per class used in the synthetic-spectra experiments. Each curve is a Lorentzian profile sampled on an $n=100$-point axis, with peak centre jittered as $c \sim \mathcal{N}(50,10^2)$. Class~0 (blue) and Class~1 (orange) differ only through the  FWHM $\xi_1=7$ vs.\ $\xi_2=9$, illustrating that the two classes are visually difficult to distinguish despite being statistically separable in high dimension.}
    \label{fig:example_spectra_only}
\end{figure}

In a second experiment S2, we then studied the classification of the spectra constituted by a single Lorentzian peak with different FWHM. For a given number of dimensions $n\in\{5,10,50,\allowbreak 100,1000,\allowbreak 2000,5000,10000\}$ we generate $N=500$ spectra per class. Each spectrum is a Lorentzian profile with a centre randomly chosen from a normal distribution with parameters $(\mu,\sigma_c)=(50,10)$ and a different FWHM $\xi_1=7$, and $\xi_2=9$ for the two classes.
For the spectroscopist, in Figure \ref{fig:example_spectra_only}, 10 examples of each class are plotted together, to show how visually it is impossible to distinguish the two classes. 

In the third experiment S3 we studied the effect of dimensionality on classification in presence of noise. In fact, in spectroscopy, for a multitude of reasons (detector dark current, electronics, etc.), noise is always present. We know from the experiments described here that it is possible to perfectly classify pure noise. It is an interesting question what happens if noise is added, for example, to a set of undistinguishable spectra (experiment S1).

For this experiment, we added to the spectra of experiment S1, an independent Gaussian noise that differs only in its mean between classes: for class~0 we add $\varepsilon^{(0)}_{j,i}\sim\mathcal N(0,0.01^2)$, and for class~1 $\varepsilon^{(1)}_{j,i}\sim\mathcal N(0.01,0.01^2)$. 
We consider $N=500$ spectra per class and $n\in\{5,10,50,100,1000,2000,5000\}$.

For each $n$ we create a balanced dataset and benchmark four standard classifiers: logistic regression (max\_iter=3000), $k$–nearest neighbours, a decision tree (max\_depth=5), and a random forest (100 trees, fixed seed). Performance is estimated using a 5-fold stratified cross-validation. We report the mean and standard deviation of validation accuracy across folds. All random draws use fixed pseudorandom seeds to ensure reproducibility.
Because peak shape and centre statistics are identical across classes, the only class–dependent signal arises from a tiny global shift in the additive noise mean. This setting reflects common practice where models ingest spectra without explicit peak annotations; the experiment investigates how classifiers exploit minute distributional offsets when presented with high–dimensional spectral vectors.

\subsection{Real Spectroscopic Data}
\label{sec:methods_real_data}

The dataset used in this study consists of fluorescence spectra of Spanish olive oil acquired with a miniaturised and low-cost fluorescence-based instrument by the authors~\cite{venturini2021exploration}. 
The dataset consists of 24
olive oil samples from the 2019--2020 harvest. The samples were classified by the producer, Conde de Benalúa (Granada, Spain) into 12 extra virgin oils (EVOO), 8 virgin oils (VOO), and 7 lampante oils (LOO), based on standard chemical and sensory parameters according to EU regulations.
Each olive oil sample was measured undiluted and under ambient conditions.  
We refer the reader to the original article \cite{venturini2021exploration} for more details. Note that the spectra used in this work have not been normalised and have been used raw.
\begin{figure}[t!]
    \centering
    \includegraphics[width=0.75\linewidth]{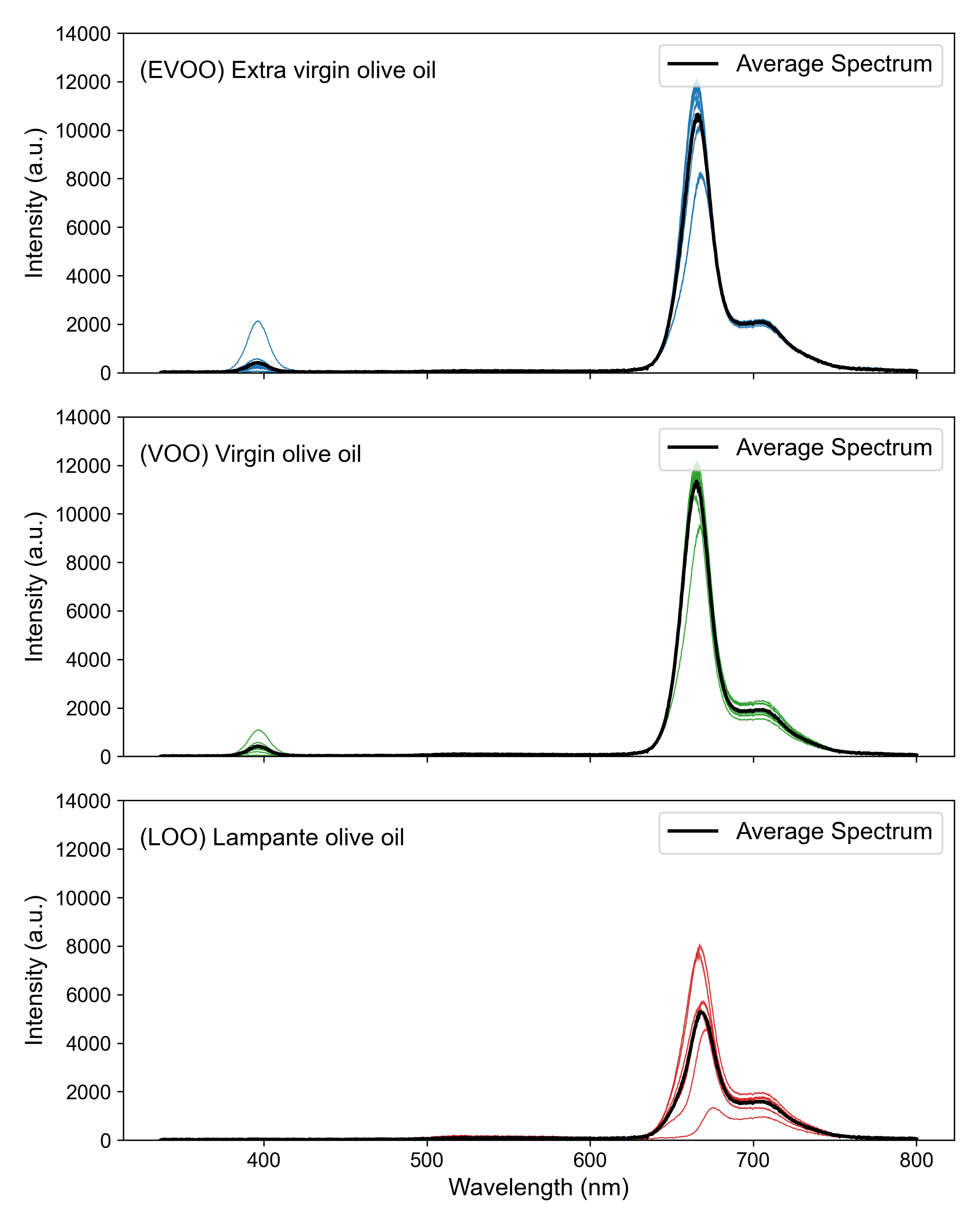}
    \caption{Fluorescence spectra of Spanish olive oil samples classified as Extra Virgin (EVOO), Virgin (VOO), and Lampante (LOO). The region 380--420 nm indicates the Rayleigh scattering peak from the excitation LED. The black line indicates the average spectrum for each class.}
    \label{fig:spectra}
\end{figure}
Figure \ref{fig:spectra} shows the spectra for the three available classes: EVOO, VOO, and LOO. In the raw data the Rayleigh scattering peak due to the excitation LED is clearly visible for EVOO and VOO but almost inexistent for LOO due to an increase absorbance in the blue-ultraviolet. Its presence, therefore, could allow models to easily distinguish, for example, EVOO from LOO. Since this work focuses on the effect of high dimensionality and minimal statistical differences on model classification performance, a spectral region between 380 and 420 nm around the excitation LED wavelength (395 nm) was eliminated.
Furthermore, two classification experiments of different complexity were performed: EVOO vs. LOO, with clearly different spectral characteristics, and the less easily distinguishable EVOO vs. VOO. All experiments of the first type are indicated with Ra1, Ra2, etc., while those of the second type Rb1, Rb2, etc. The experiments performed with these data are summarised in Table \ref{tab:exp-overview}.



\section{Results}
\label{sec:results}

This section presents the results of the experiments. For clarity, the overview of the results for each experiment can be found in Table \ref{tab:exp-findings}.

\begin{table}[hbt]
\caption{Main findings and references to results.}
\label{tab:exp-findings}
\centering\footnotesize
\setlength{\tabcolsep}{4pt}
\begin{tabular}{@{}p{.05\textwidth}p{.22\textwidth}p{.40\textwidth}p{.30\textwidth}@{}}
\toprule
\textbf{ID} & \textbf{Experiment} & \textbf{Key finding} & \textbf{Reference to Results} \\
\midrule \midrule

N1 & Gaussian noise: $\Delta\sigma$ sweep (QDA)
& Accuracy climbs monotonically with $\Delta\sigma$ and with dimension $n$; it reaches almost 1 already with modest gaps for $n$ high enough. In high $n$ white noise is easily and perfectly classifiable. We have tested $n\in \{ 5,10,50,500 \}$ and $\Delta \sigma$ from 0 to 2 for an homogoenous covariance and for a Toeplitz one with $\rho=0.95$.
& Fig.~\ref{fig:accuracy_vs_sigma_difference}; Methods in Section \ref{sec:gauss1}; Results in Section \ref{sec:results_gaussian_noise} \\

\midrule
N2 & Gaussian noise: Bayes “oracle” boundary
& White noise from the two distributions is almost perfectly classifiable in high enough $n$; accuracy goes to 1 quickly as $\Delta\sigma$ grows. We have tested $n\in \{ 30, 100, 500, 1000, 5000 \}$ and we have varied $\Delta \sigma$ from 0 to 1.0.
& Fig.~\ref{fig:accuracy_vs_sigma_difference_bayes1}; Methods in Section \ref{sec:gauss1}; Results in Section \ref{sec:results_gaussian_noise} \\

\midrule
N3 & Gaussian noise: accuracy vs.\ dimension $n$
& Even small $\Delta\sigma$ becomes highly separable as $n$ increases (dimensionality amplifies tiny distributional gaps). We have tested $n$ from 0 to 5000 (but visualisaed only until 100, to make the more steeper growing curves more visible) and tested $\Delta \sigma \in \{ 0.1, 0.3, 0.6, 0.9, 1.2, 1.5, 2.0 \}$.
& Fig.~\ref{fig:accuracy_vs_sigma_difference2}; Methods in Section \ref{sec:gauss1}; Results in Section \ref{sec:results_gaussian_noise} \\

\midrule
N4 & Skew–normal noise: 2D parameter sweeps
& Tiny shifts in mean, variance or skewness give near-perfect accuracy for most models; random forest saturates the fastest. 
& Fig.~\ref{fig:acc-vs-pars-skew}; Methods in Section \ref{sec:skew}; Results in Section \ref{sec:results_skew}; Results in Section \ref{sec:synth-spectra} \\

\midrule
S1 & Synthetic spectra: truly identical classes
& No classifier exceeds chance level accuracy (0.5) confirming that without distributional differences, classification is impossible.
& Tab.~\ref{tab:undi1}; examples in Fig.~\ref{fig:example_spectra_only} \\

\midrule
S2 & Synthetic spectra: width difference 
& Validation accuracy rises as expected with $n$; linear and ensemble models approach $1.0$ for large $n$ similarity in the spectra.
& Fig.~\ref{fig:model-accuracy-vs-n}; Results in Section \ref{sec:synth-spectra} \\

\midrule
S3 & Synthetic spectra with additive noise offset
& High $n$ allows a minute noise distributional difference to enable near-perfect separability; random forest reaches approximately 1.0 with very small values of $n$.
& Fig.~\ref{fig:model-accuracy-vs-n_noise-offset}; Results in Section \ref{sec:synth-spectra} \\

\midrule
Ra1/Rb1 & Global pixel permutation 
& Accuracy remains high ($\sim 82\%$) despite the total destruction of spectral shapes. This empirically proves the model relies on global covariance structures rather than physical spectroscopic peaks.
& Results in Sec.~\ref{sec:oil} \\

\midrule
Ra2/Rb2 & Independent row permutation 
& Model performance collapses to the majority-class baseline. This confirms that success in Ra3/Rb3 was due to high-dimensional statistical structure, which is destroyed by independent shuffling.
& Results in Sec.~\ref{sec:oil} \\

\midrule
Ra3/Rb3 & Pixel count sweep 
& Accuracy reaches $>85\%$ using only 15--20 randomly selected pixels from the noise region ($\rho_1$). This confirms that non-contiguous, chemically empty data provide sufficient statistical separation in high dimensions.
& Fig.~\ref{fig:random_subsets}; Results in Sec.~\ref{sec:oil} \\

\midrule
Ra4/Rb4 & Feature importance: window sweep 
& High classification accuracy ($\sim 80\%$) is maintained across all windows, including those in the signal-free region ($0$--$400$ px). Larger windows ($W=400$) create an accuracy plateau independent of spectral features.
& Fig.~\ref{fig:sliding_window_comparison}; Results in Sec.~\ref{sec:oil} \\

\midrule
Ra5/Rb5 & Feature importance: SHAP
& SHAP attribution is distributed across the entire spectrum, often assigning higher "importance" to noise regions than to chemical peaks. This highlights the "interpretability paradox" in high dimensions.
& Fig.~\ref{fig:shap}; Results in Sec.~\ref{sec:oil} \\



\bottomrule
\end{tabular}
\end{table}

\subsection{Gaussian Noise Classification}
\label{sec:results_gaussian_noise}

\begin{figure}
    \centering
    \includegraphics[width=0.75\linewidth]{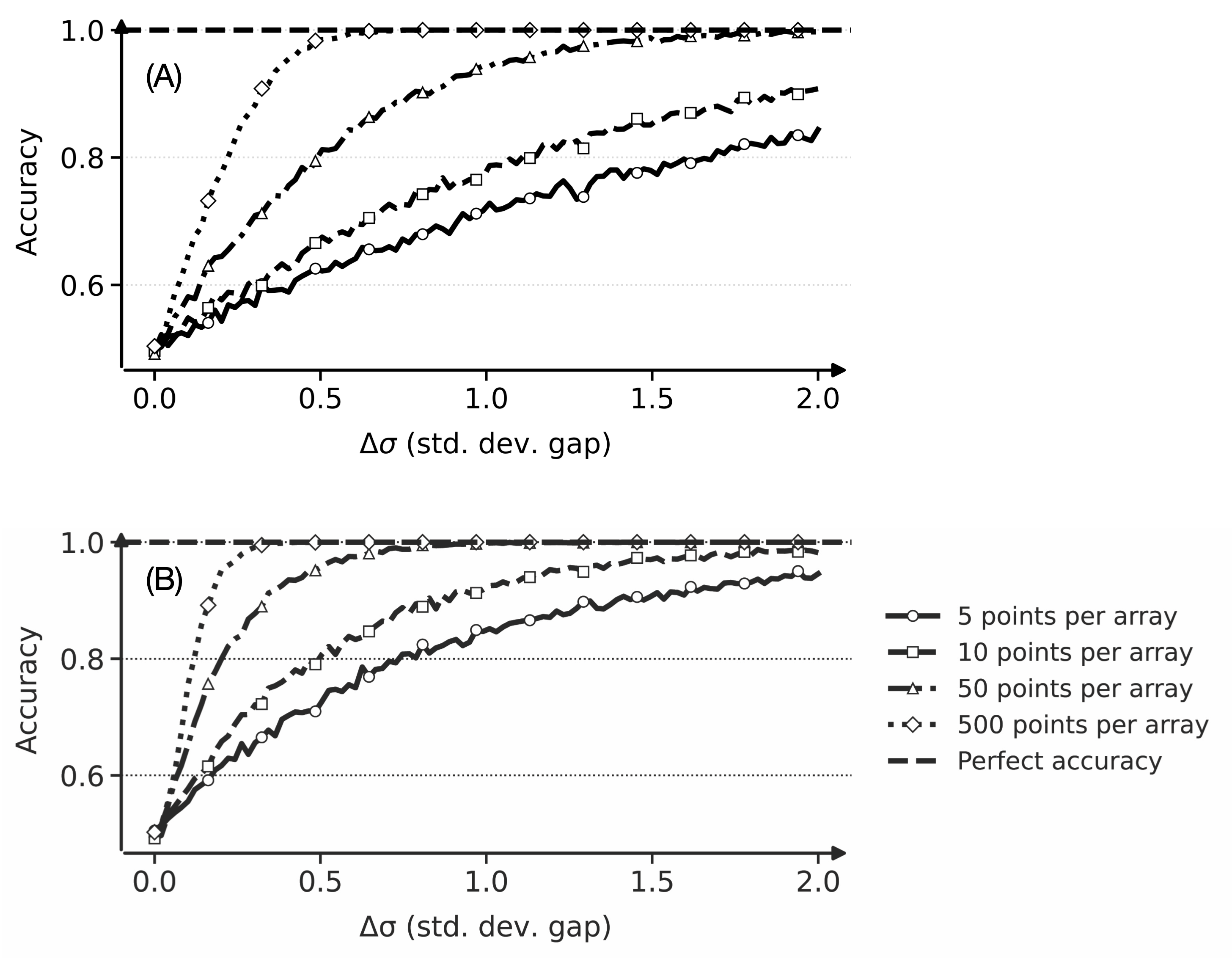}
    \caption{Results from experiment N1. Classification accuracy of QDA (regularisation parameter equal to 0.4) as a function of the standard-deviation gap $\Delta\sigma$ between two white-noise classes with equal mean $\mu=1$ and baseline $\sigma_1=1$. Each curve corresponds to a different number of points per array ($n\in{[5,10,50,500]}$); at each $\Delta\sigma$, $N$ arrays per class are generated and split 80/20 into train/test. The dashed line at $1.0$ marks perfect accuracy. The results are for the test dataset. Panel (A) has been obtained with a Toeptlitz covariance with $\rho=0.95$, while panel (B) with a homogeneous covariance. It is an interesting observation that adding correlations between neighbouring values, slow down the growth to perfect accuracy, but it does not stop it altogether. }
    \label{fig:accuracy_vs_sigma_difference}
\end{figure}

The first experiment (N1) consisted of classifying two classes of $n$-dimensional noise arrays from isotropic and for Toeplitz covariances:
$\mathcal{N}(\mu_1,\sigma_1^2 I_n)$ and $\mathcal{N}(\mu_2,\sigma_2^2 I_n)$ and for $\mathcal{N}(\mu_1,\Sigma_1(\rho))$ and $\mathcal{N}(\mu_2,\Sigma_1(\rho))$, respectively.
The results of the quadratic discriminant analysis (QDA) with a regularisation parameter equal to 0.4 are shown in Figure \ref{fig:accuracy_vs_sigma_difference}. In this simulation, we fix $\mu_1=\mu_2$ and sweep the standard–deviation gap $\Delta\sigma:=|\sigma_2-\sigma_1|$ over $[0,2]$, while varying the dimensionality $n$ (points per array). For each $\Delta\sigma$ we formed an 80/20 train–test split and reported the test accuracy. The latter is close to chance ($\approx 0.5$) when $\Delta\sigma$ is close to zero and increases with both $\Delta\sigma$ and $n$; higher dimensions reach approximately $1.0$ rather quickly with much smaller gaps (e.g., hundreds of points per array achieve near-perfect accuracy for modest $\Delta\sigma$), whereas very small $n$ require larger gaps to exceed $0.9$. 
Panel (A) in Figure \ref{fig:accuracy_vs_sigma_difference} shows the results for a Toeplitz covariance with $\rho=0.9$ and panel (B) for a homogeneous one. Notably, when considering Toeplitz matrices, the presence of correlation slows down the effect of high dimensionality (effectively it takes large gaps $\Delta \sigma$ for the same $n$ value to reach the same accuracy value), but it does not stop it (as is expected, since the Feldman-H\'ajek theorem is valid for generic covariances).

\begin{figure}[hbt]
    \centering
    \includegraphics[width=.75\linewidth]{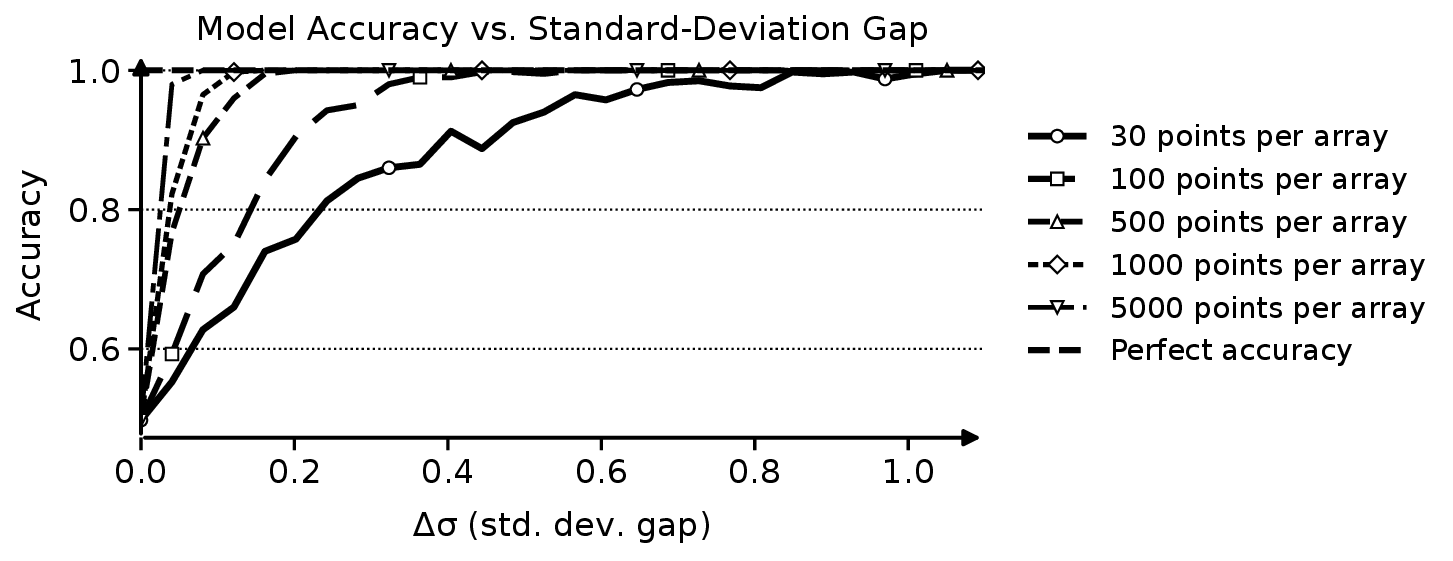}
    \caption{Results from experiment N2. Accuracy of the Bayes classifier for two Gaussian white-noise classes with common mean $\mu=10$ and variances $\sigma_1^2 I_n$ vs.\ $\sigma_2^2 I_n$. The decision uses the sufficient statistic $S=\sum_{j=1}^{n}(x_j-\mu)^2$ with the LDA threshold $T$.}
    \label{fig:accuracy_vs_sigma_difference_bayes1}
\end{figure}

In the second experiment (N2) for each variance gap value $\Delta\sigma$, we generated two classes of $N=1000$ arrays for various values of $n$ from isotropic Gaussians with equal mean $\mu$ and standard deviations $\sigma_1$ and $\sigma_2=\sigma_1+\Delta\sigma$.
We evaluated the LDA decision boundary $T$ for this setting
\begin{equation}
\label{eq:T}
    T=n\log \left(\frac{\sigma_2^2}{\sigma_1^2}\right) \frac{1}{\displaystyle\frac{1}{\sigma_2^2}-\frac{1}{\sigma_1^2} }
\end{equation}
which classifies a sample by thresholding its squared norm around this $T$ value. For each $\Delta\sigma$ we formed an 80/20 train–test split and computed the test
accuracy on this Bayes decision boundary. The results can be seen in Figure \ref{fig:accuracy_vs_sigma_difference_bayes1}. When
$\Delta\sigma=0$ ($\sigma_1=\sigma_2$) the classes are indistinguishable and the
accuracy is approximately $0.5$; as $\Delta\sigma$ grows, the distributions become
increasingly separable and the accuracy approaches $1$, matching the theoretical
behaviour for isotropic Gaussians. Note that if we consider Toeplitz covariances, the same \textit{slowing down} effect described for Figure \ref{fig:accuracy_vs_sigma_difference} appears. We have not reported these additional results to keep the length of this article reasonable.

\begin{figure}[hbt]
    \centering
    \includegraphics[width=.75\linewidth]{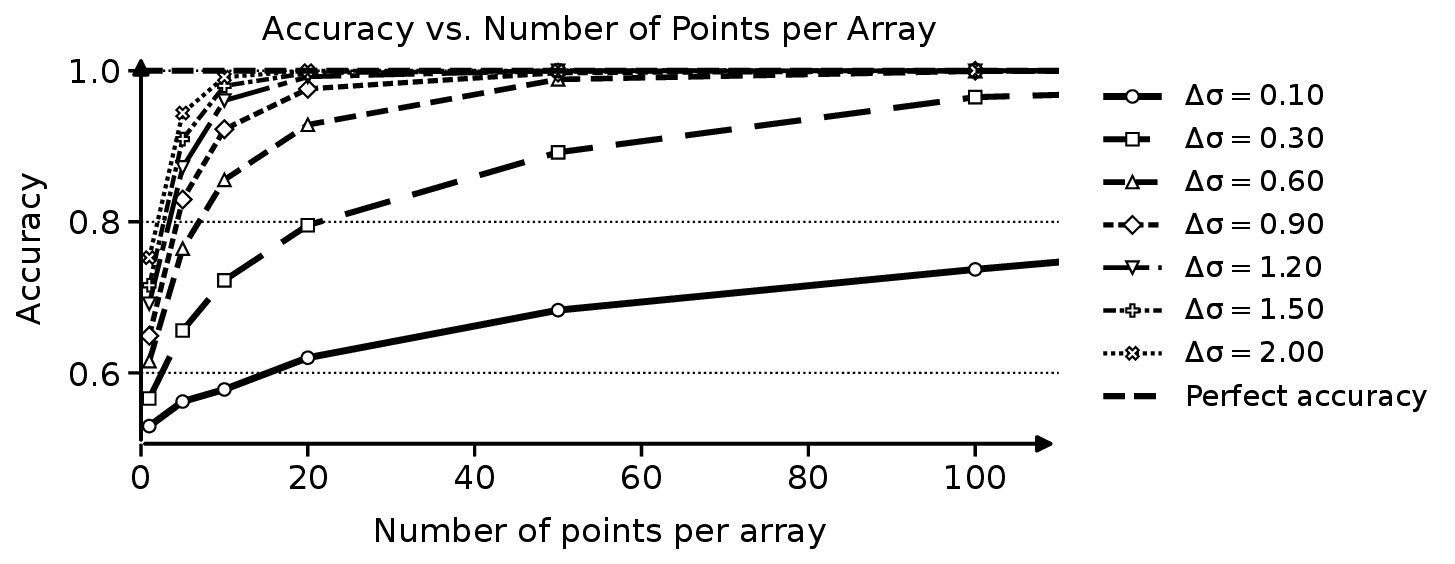}
    \caption{Results from experiment N3. Test accuracy of QDA (reg\_param $=0.4$) versus the number of points per array $n$ for two Gaussian white-noise classes with common mean $\mu$ and variances $\sigma_1^2 I_n$ vs.\ $\sigma_2^2 I_n$. Each curve corresponds to a different standard-deviation gap $\Delta\sigma=\sigma_2-\sigma_1$. Datasets contain $N$ arrays per class and are split $80/20$ into train/test. The dashed horizontal line at $1.0$ indicates perfect accuracy. Accuracy increases with both the number of points per array $n$ and the variance gap $\Delta\sigma$; even small $\Delta\sigma$ yields near-perfect accuracy as $n$ grows—showcasing how high dimensionality amplifies differences in distributions.}
    \label{fig:accuracy_vs_sigma_difference2}
\end{figure}

In the third experiment (N3) we generated two classes of
$N=1000$ arrays from isotropic Gaussians with equal mean $\mu=0$ and standard
deviations $\sigma_1=1$ and $\sigma_2=\sigma_1+\Delta\sigma$. The results are shown in Figure \ref{fig:accuracy_vs_sigma_difference2}, for each standard deviation gap $\Delta \sigma \in \{0.1, 0.3, 0.6, 0.9, 1.2, 1.5, 2.0\}$ and dimension
$n \in \{1, 5, 10, ,\allowbreak20, 50, \allowbreak 100, 200,\allowbreak 500, 1000, 2000,\allowbreak 5000\}$. We used QDA
and calculated the test accuracy on a $20\%$ hold-out split for each $(n,\Delta\sigma)$.
The curves show that accuracy increases monotonically with $n$ and with the gap
$\Delta\sigma$: larger gaps reach near-perfect accuracy at much smaller $n$,
while very small gaps require higher dimensions to move far above chance.

\subsection{Skew Normal Noise Classification}
\label{sec:results_skew}

\begin{figure*}
    \centering
    \includegraphics[width=0.9\linewidth]{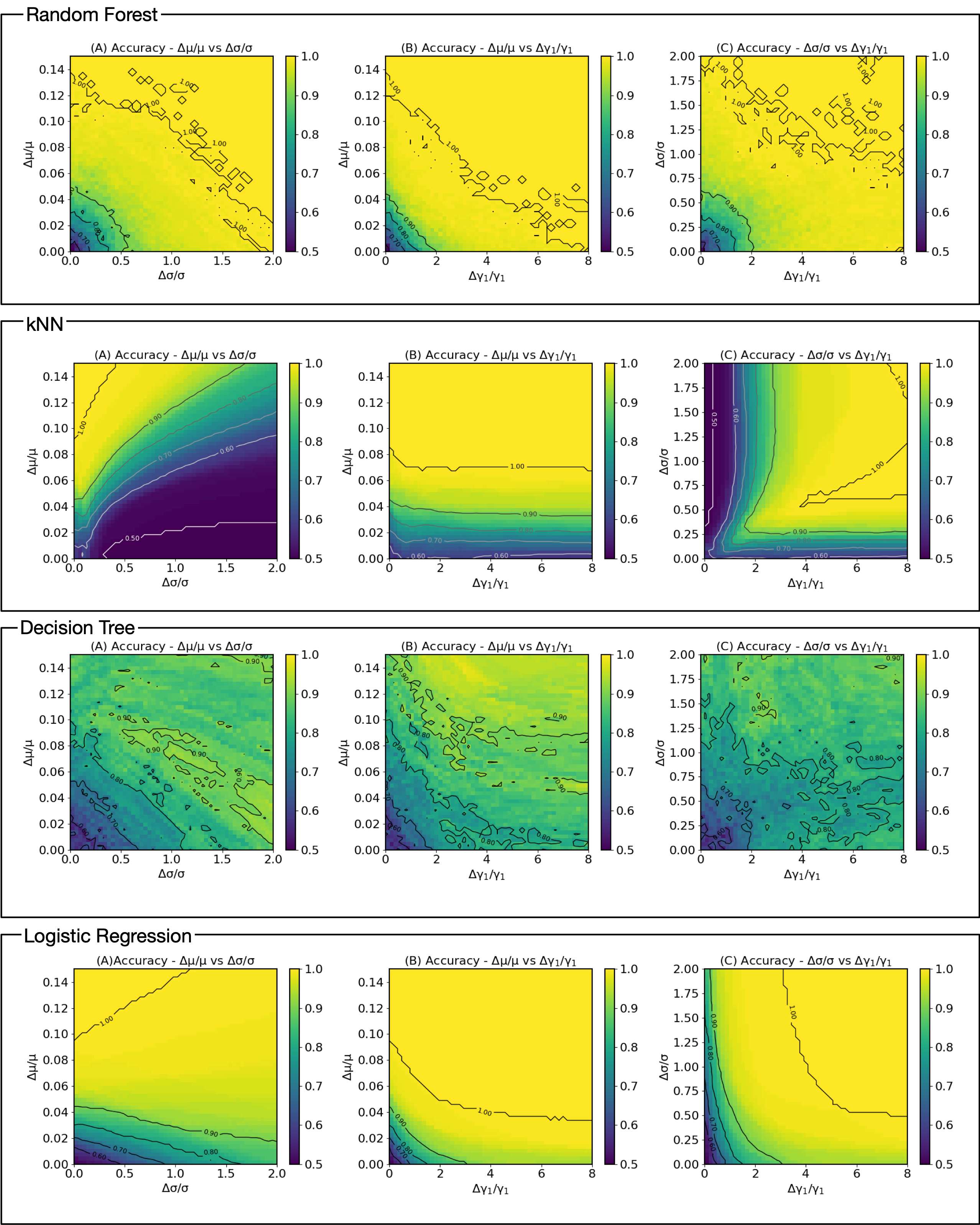}
    \caption{Results from experiment N4. Cross‐validated classification accuracy for four models (from top to bottom: Random Forest, kNN, Decision Tree, Logistic Regression) on synthetic data drawn from a skew–normal distribution in dimension $n=50$. Each column sweeps two parameters while holding the third at its baseline: (A) $\Delta\mu/\mu_1$ vs.\ $\Delta\sigma/\sigma_1$, (B) $\Delta\mu/\mu_1$ vs.\ $\Delta\gamma/\gamma_1$, (C) $\Delta\sigma/\sigma_1$ vs.\ $\Delta\gamma/\gamma_1$, where $\gamma$ is the skew (shape) parameter of the skew–normal. Class~0 is fixed at $(\mu_1,\sigma_1,\gamma_1)=(10,1,0.5)$ and Class~1 at $(\mu_2,\sigma_2,\gamma_2)=(\mu_1+\Delta\mu,\ \sigma_1+\Delta\sigma,\ \gamma_1+\Delta\gamma)$. For every grid point we generate $N=100$ samples per class and report the mean accuracy over $5$‐fold cross validation. Colour encodes accuracy (scale at right, 0.5–1.0); black contour lines mark iso‐accuracy levels.}
    \label{fig:acc-vs-pars-skew}
\end{figure*}

The results of experiment N4 are shown in Figure~\ref{fig:acc-vs-pars-skew}, where the cross–validated accuracy for four models trained to distinguish two classes drawn from a skew–normal law in dimension $n=50$.  
Class~0 is fixed at $(\mu_1,\sigma_1,\gamma_1)=(10,1,0.5)$ and Class~1 at $(\mu_2,\sigma_2,\gamma_2)=(\mu_1+\Delta\mu,\ \sigma_1+\Delta\sigma,\ \gamma_1+\Delta\gamma)$.  
Each plot sweeps two parameters while the third remains at its baseline: (A) $\Delta\mu/\mu_1$ vs. $\Delta\sigma/\sigma_1$, (B) $\Delta\mu/\mu_1$ vs.\ $\Delta\gamma/\gamma_1$, (C) $\Delta\sigma/\sigma_1$ vs.\ $\Delta\gamma/\gamma_1$.  
Contour lines help identify exact accuracy values. The colour scale ranges from chance (0.5) to perfect (1.0).

Figure~\ref{fig:acc-vs-pars-skew} shows that logistic regression achieves a near-perfect performance in most parameter ranges. A small mean shift ($\Delta\mu$) already drives the accuracy to $\approx 1.0$, and combinations of variance and skew differences ($( \Delta\sigma,\Delta\gamma)$) also reach 100\% accuracy very quickly. 
Random forest shows very strong and robust results across the three plots, with broad yellow (1.0) regions, and it saturates quickly at perfect accuracy. kNN requires larger separations to leave chance accuracy and shows the steepest transition bands. This is consistent with the curse of dimensionality: local neighbourhoods are less informative in $n=50$, so $k$NN needs larger $(\Delta\mu,\Delta\sigma,\Delta\gamma)$ to separate the classes. In contrast, decision tree shows intermediate performance with more granular contours. Random forest exhibits the highest accuracy even for very small differences in the parameters.

To summarise, already in $n=50$ dimensions, which is low compared to typical spectrum dimensions (of the order of $10^3$), tiny discrepancies in mean, spread, or skewness already make the classes almost perfectly distinguishable. Figure ~\ref{fig:acc-vs-pars-skew} shows that all four models approach the $100\%$ accuracy in wide regions of the parameter grids.

\subsection{Synthetic Spectra Classification}
\label{sec:synth-spectra}



In experiment S1, as expected, no model is able to distinguish between the two classes, as the results in Table~\ref{tab:undi1} show. This outcome confirms that when the underlying data distributions are truly identical, the classification task becomes statistically impossible: all models perform at chance level, with validation accuracies fluctuating around 0.5 due to random sampling. This serves as a sanity check, demonstrating that the experimental setup and models behave consistently with theoretical expectations.
\begin{table}[htb]
\centering
\caption{Accuracy of multiple models for indistinguishable spectra generated as described in the text. Small
deviations from $0.5$ are attributable to finite-sample fluctuations.}
\label{tab:undi1}
\begin{tabular}{l c}
\toprule
\textbf{Model} & \textbf{Accuracy (mean $\pm$ SD)} \\
\midrule
Logistic Regression      & $0.54 \pm 0.03$ \\
K-Neighbors Classifier   & $0.50 \pm 0.04$ \\
Decision Tree Classifier & $0.53 \pm 0.04$ \\
Random Forest Classifier & $0.51 \pm 0.01$ \\
\bottomrule
\end{tabular}
\end{table}

\begin{figure}[hbt]
    \centering
    \includegraphics[width=0.75\linewidth]{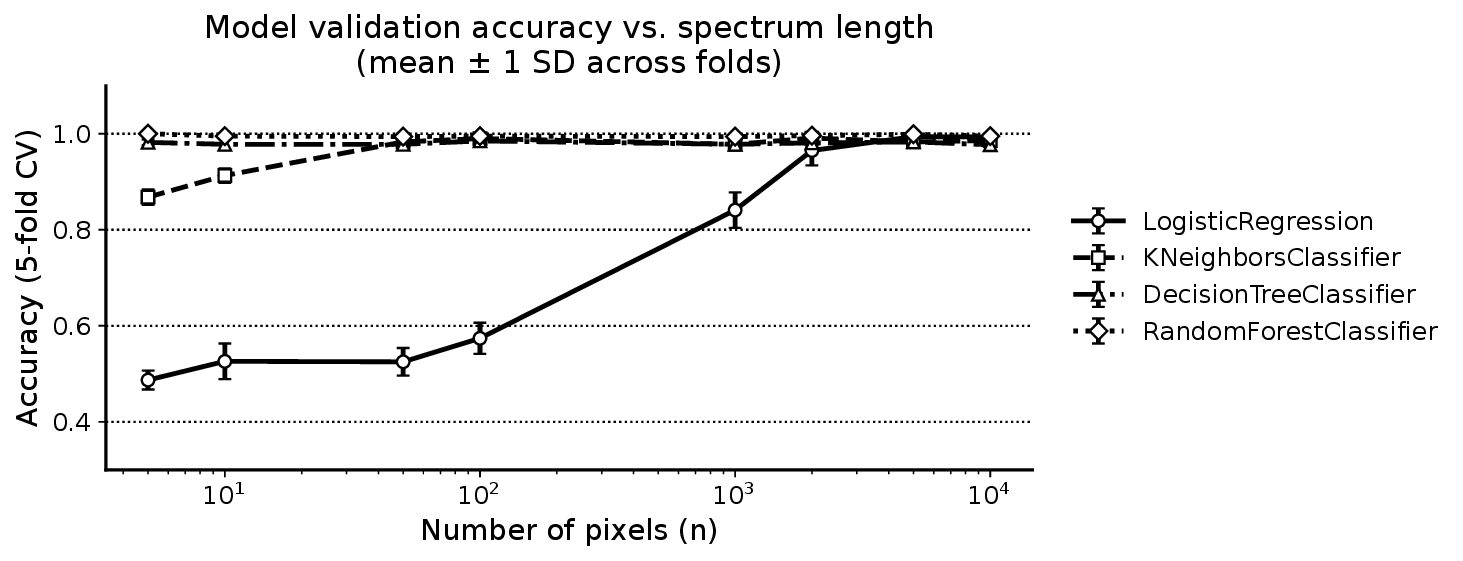}
    \caption{Results from experiment S2. Model validation accuracy (mean $\pm$1 SD over 5-fold CV) versus spectrum length $n$ (log scale)
for four classifiers: logistic regression, $k$-NN, decision tree (max depth 5), and random forest (100 trees).
Data consist of synthetic one–peak Lorentzian spectra: the two classes differ only in width
($\xi_1=7$ vs.\ $\xi_2=9$) while peak centres are jittered $c\sim\mathcal N(50,10^2)$; 
$500$ spectra per class are generated for each $n$.
Markers show the mean accuracy with error bars; the legend identifies the models. }
    \label{fig:model-accuracy-vs-n}
\end{figure}

In experiment S2, we observe that increasing the dimensionality enables even simple models such as logistic regression to achieve near-perfect accuracy. As shown in Figure~\ref{fig:model-accuracy-vs-n}, classification performance improves steadily with the number of intensity points (or dimensions) in all models tested. These results were obtained using 5-fold stratified cross-validation and averaged over folds, with all random seeds fixed for reproducibility. These findings illustrate how, in typical spectroscopic ML analysis, where spectra are treated as high-dimensional vectors, classifiers may exploit subtle distributional differences unrelated to the actual chemical structure.

In all models, validation accuracy increases with the size of the spectrum $n$, reflecting the fact that the differences in FWHM become easier to detect. Linear and ensemble methods benefit the most from increasing $n$, with performance approaching unity for large arrays, whereas shallow trees saturate earlier. Training accuracies follow the same trend, indicating low variance for the ensemble and stable generalisation once $n$ is sufficiently large.
This controlled study isolates a single physical change (peak width) under substantial positional jitter and shows how dimensionality alone can convert a weak per-pixel signal into a highly separable representation, providing a transparent explanation for model behaviour on spectroscopic data.

\begin{figure}[hbt]
    \centering
    \includegraphics[width=.75\linewidth]{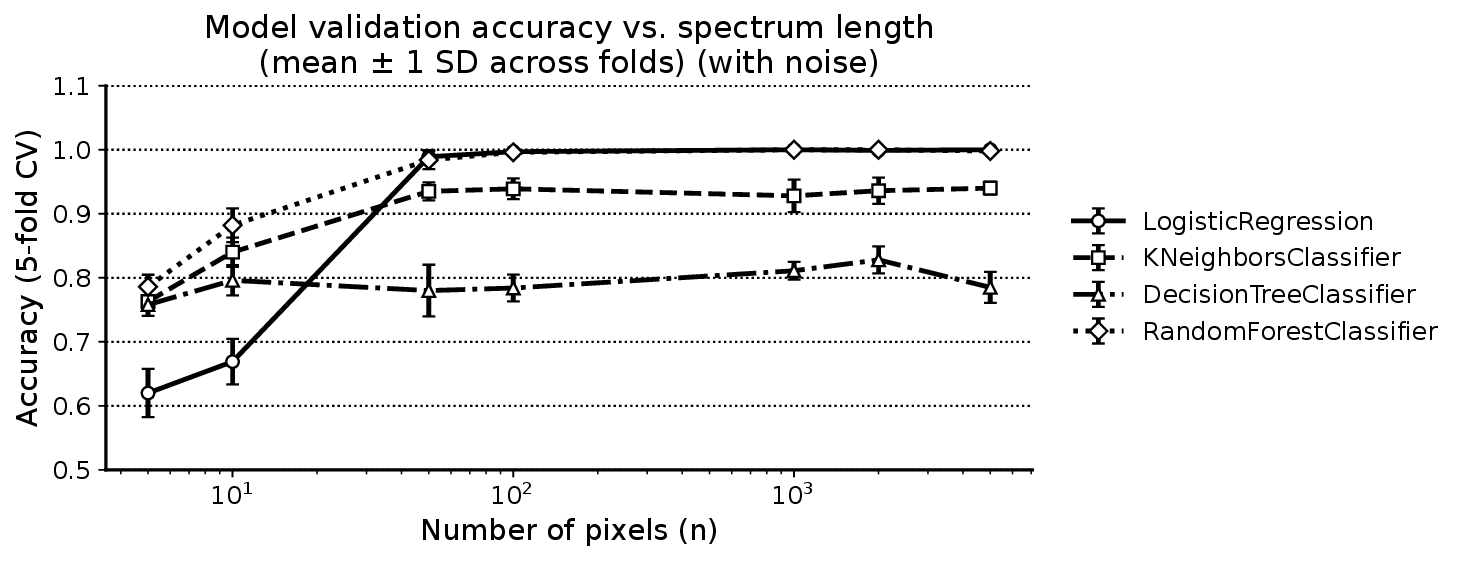}
    \caption{Results from experiment S3. Validation accuracy (mean $\pm$1~SD over 5-fold CV) versus spectrum length $n$ (log scale) for four classifiers on synthetic one–peak spectra with identical signal distributions but class-specific additive noise. 
Each class contains $500$ spectra with Lorentzian FWHM $\xi=7$ and centres jittered $c\sim\mathcal N(50,10^2)$; noise is i.i.d.\ Gaussian with mean $0$ (class~0) or $0.01$ (class~1) and SD $0.01$. 
Accuracy rises with $n$ for all methods: random forests reach $\approx\!1.0$ with tens of intensity points, $k$-NN stabilizes around $0.93$–$0.95$, logistic regression climbs steadily toward $1.0$, while the single decision tree plateaus near $0.8$.}
    \label{fig:model-accuracy-vs-n_noise-offset}
\end{figure}

The results of experiment S3 are shown in Figure \ref{fig:model-accuracy-vs-n_noise-offset}.
The two classes differ only by a small class–specific offset in additive noise (mean $0$ vs.\ $0.01$ with SD $0.01$); the underlying Lorentzian signal (centre distribution and width) is identical. The figure illustrates a general phenomenon in spectroscopy: when models have as input spectra as high–dimensional vectors, even subtle distributional shifts (here, a $0.01$ mean offset in the noise) can become highly separable as $n$ increases. The ensemble and nearest-neighbour
methods aggregate this diffuse evidence quickly; linear models eventually catch up as the $\sqrt{n}$ gain in averaging overwhelms noise; shallow single trees remain bias–limited.

    
\subsection{Real Spectroscopic Data Classification}
\label{sec:oil}

In this section, we show this phenomenon with data from real spectroscopic measurements.

\subsubsection{Spectral Regions and Statistical Differences}

Let us discuss the main five different regions in the spectra.
\begin{itemize}
    \item \textbf{Region $\rho_1$:} 337 nm -- 380 nm: this region contains only noise and no chemical fingerprint.
    \item \textbf{Region $\rho_2$:} 380 nm -- 420 nm: this region contains the Rayleigh scattering peak. As explained, this region has been removed from the spectra, to take away an easy way to a high accuracy for models.
    \item \textbf{Region $\rho_3$:} 420 nm -- 630 nm: this region contains very weak fluorescence signals, and as such chemical information, mainly due to the oxydation products in olive oil.
    \item \textbf{Region $\rho_4$:} 630 nm -- 775 nm: this region contains the strongest fluorescence signals (due to chlorophylls) and corresponds to the strongest chemical information.
    \item \textbf{Region $\rho_5$:} 775 nm -- 800 nm: this region, similar to region $\rho_3$, contains only weak chemical information, since only the tails of the main peak are present in this region.
    
\end{itemize}

To study the effect of noise and dimensionality, we will use mainly regions $\rho_1$ and $\rho_3$. Region $\rho_4$ is less interesting for the discussion in this paper.



\subsubsection{Baseline and Classification Accuracy with the Entire Spectra}

First of all, it is important to establish a baseline. The easiest model we can think of is a majority class classified (classifying every sample in the majority class). Doing this will give us an accuracy of 63\% for EVOO vs. LOO, and 60\% for EVOO vs. VOO. Doing a classification of the spectra with a random forest model (with 100 estimators) gives us a Leave-one-out cross validation (LOO-CV) accuracy of 90\% and 80\%, respectively.

\subsubsection{Global Pixel Permutation}

To rigorously test whether the high accuracy is driven by the spatial arrangement of intensities (spectroscopic features) or merely by high-dimensional statistical geometry (as we claim), we performed a global pixel permutation experiment (experiment Ra1/Rb1). We applied a single, identical for all spectra random shuffle to all pixels across the entire dataset. This process preserves the statistical properties of the classes (mean and covariance) while completely destroying all physical context, such as peaks, baselines, and continuity.

Remarkably, a random forest classifier trained on these ``scrambled'' spectra achieved an accuracy of 82\% for EVOO vs. LOO and 81\% for EVOO vs. VOO. Since a peak cannot exist in a shuffled vector, this result serves as evidence: the model is not ``reading'' the spectra in any chemical sense. Instead, it is exploiting the concentration of measure in high-dimensional space. This confirms that in $10^3$ dimensions, class-specific noise patterns and instrumental offsets become perfectly separable regardless of their physical meaning. Formulated more cautiously: the noise (intended as a non fluorescence signal) provides a higher degree of statistical separability in high-dimensional space.

This global pixel permutation experiment suggests a more cautious but powerful formulation of the high-dimensionality paradox: statistical artefacts are often `easier' for models to exploit than chemical signals. Because instrumental noise and baseline offsets provide a consistent, high-dimensional footprint, flexible models (like random forests) can reach high accuracy by following the path of maximum statistical separation, even when the physical 'structure' of the data has been completely destroyed.

\subsubsection{Independent Raw Permutation}

To isolate whether the model relies on the global statistical structure or merely on individual sample intensities, we performed the experiment implemented an independent row permutation (Experiment Ra2/Rb2). In this case, each spectrum was shuffled using a unique random seed, thereby destroying the inter-pixel covariance.

Unlike the global shuffle (which yielded $\sim$82\% accuracy), this independent shuffle caused the model performance to collapse to the baseline of the majority-class ($\sim$60\%). This contrast provides a definitive multi-stage proof: the observed high accuracies in spectroscopic ML are primarily driven by the \textit{covariance structure} of non-chemical artefacts in high-dimensional space. When this structure is preserved (global shuffle), the model succeeds without chemistry; when it is destroyed (independent shuffle), the model's ``infinite-dimensional'' advantage vanishes. This confirms that the Feldman-Hájek effect, governed by class-specific covariance differences, is the functional root cause of the reported high performances.

It is important to distinguish between the preservation of statistical correlation and the preservation of chemical information. A global permutation destroys all physical contiguity and spectral shapes (the ``chemistry''), yet it leaves the underlying covariance matrix $\Sigma$ intact (although reindexed). The fact that accuracy remains at 82\% after a global shuffle proves that the model is performing a geometric separation based on these re-indexed statistical correlations rather than any recognisable spectroscopic features.

\subsubsection{Pixel Count Sweep}

The experiment R3a/R3b shows how noise in conjunction with high-dimensionality can allow a model to classify the samples.
Using randomly selected subsets of intensity values from the spectral region $\rho_1$, previously identified as containing no chemical fingerprints, together with a random forest classifier (100 estimators), we evaluated model accuracy using a LOO-CV approach. To do so, we gradually increased the number of randomly selected pixels $k$ drawn from the first 50 pixels, from 2 to 35 in both the EVOO vs. LOO and the EVOO vs. VOO classification. For each value of $k$, 20 independent random subsets were chosen.

\begin{figure}[hbt]
    \centering
    \includegraphics[width=.75\linewidth]{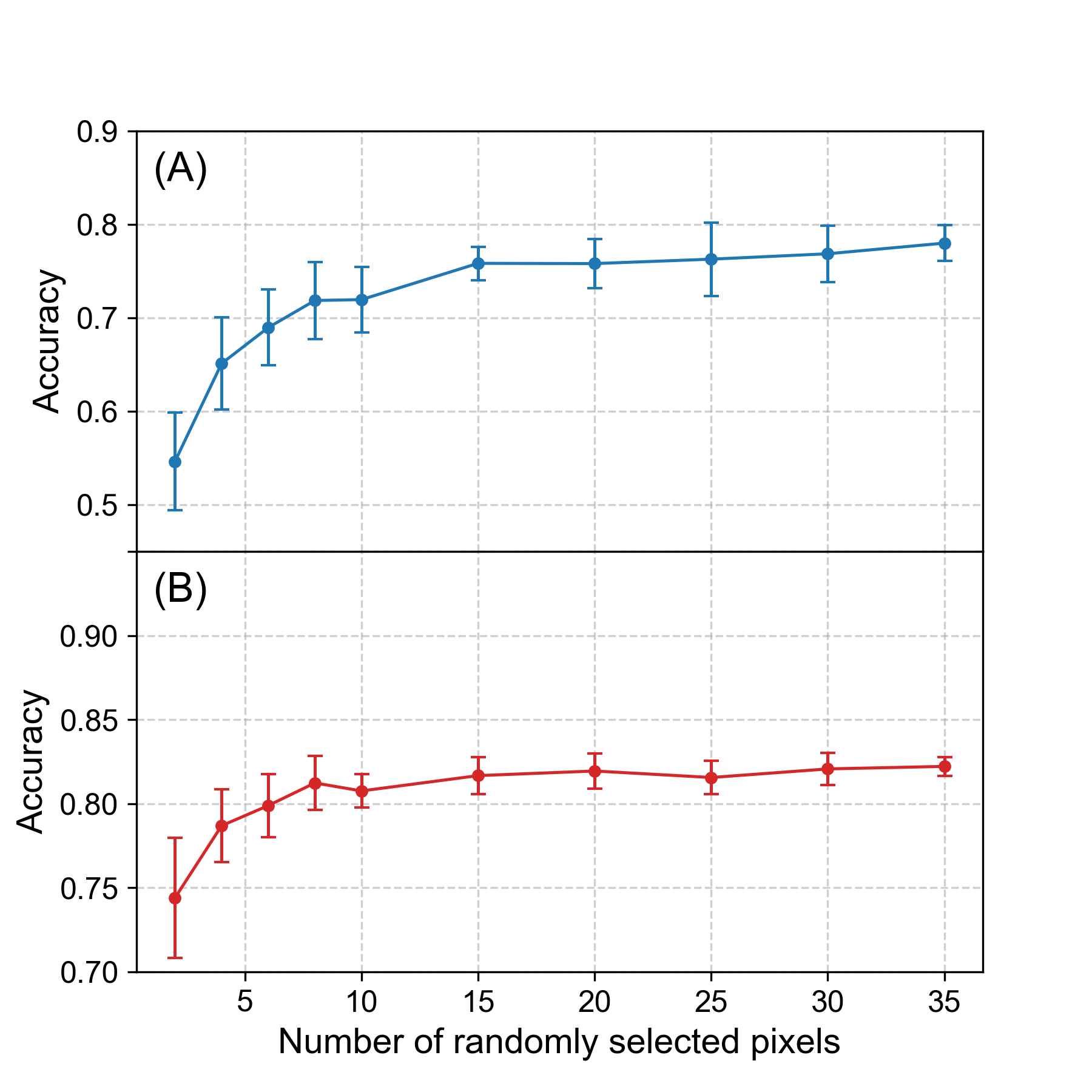}
    \caption{Results for experiment Ra3/Rb3. LOO-CV classification accuracy as a function of the number of randomly selected pixels ($k$) from the spectral noise region (pixels 0--50). Panel (A) shows the results for EVOO vs. LOO, and Panel (B) for EVOO vs. VOO. Each data point represents the mean accuracy over 20 independent random subsets, with error bars indicating the standard deviation. The rapid climb to accuracies above 80\%-90\% using only a handful of non-contiguous, chemically empty pixels serves as evidence: the model success is driven by high-dimensional statistical separability of instrumental artefacts rather than spectroscopic chemical features.}\label{fig:random_subsets}
\end{figure}

The results, shown in Figure \ref{fig:random_subsets}, provide evidence for our thesis. As the number of randomly selected pixels increases, the classification accuracy climbs steeply, reaching approximately very high values already with ca. 15-20 pixels. This occurs despite the fact that:
\begin{enumerate}
    \item The pixels are chosen from a region lacking any known physico-chemical signals.
    \item The pixels in each subset are not necessarily contiguous, destroying any potential ``hidden'' spectral shapes or features.
    \item The LOO-CV validation ensures that the model is generalising to unseen oil samples, rather than overfitting specific measurements.
\end{enumerate}

These findings suggest that the high accuracy often reported in the application of ML in spectroscopic is not necessarily a result of the model identifying complex chemical patterns. Instead, it highlights a fundamental property of high-dimensional spaces: as the dimensionality $n$ (here represented by $k$) increases, even infinitesimal distributional differences in noise or instrumental offsets between classes become almost surely separable. This experiment reinforces our cautious formulation that non-chemical noise is statistically ``easier'' for models to exploit than the subtle chemical signatures sought by researchers.

\begin{figure*}
    \centering
    \includegraphics[width=1\linewidth]{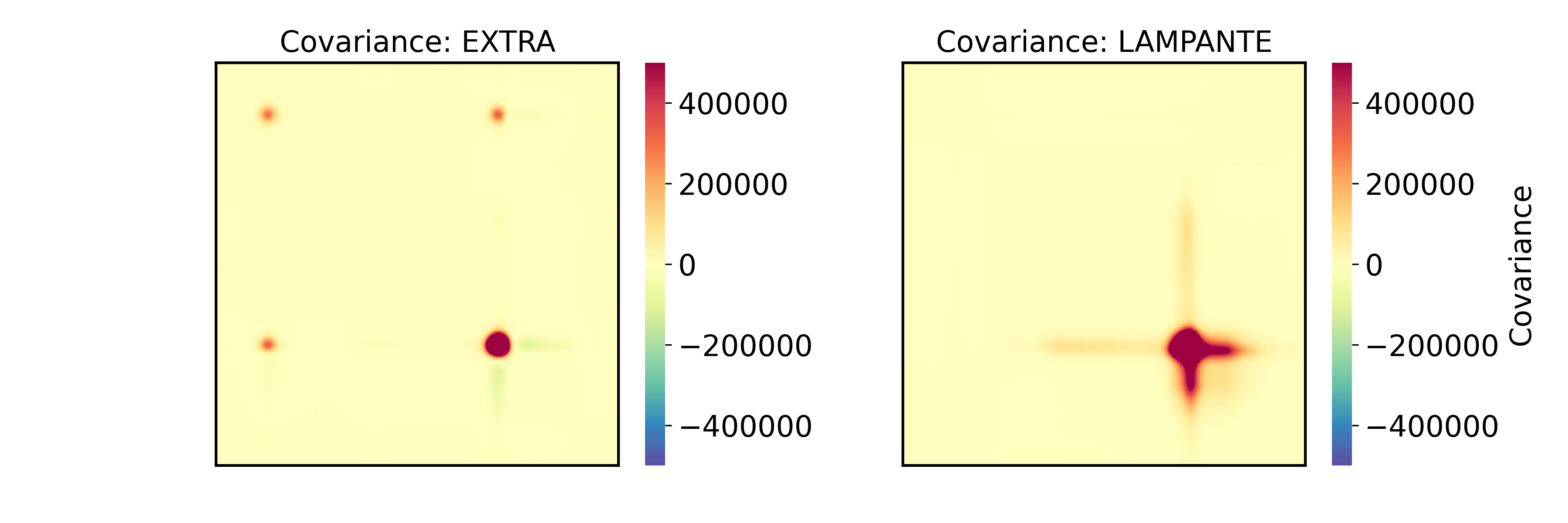}
    \caption{Empirical covariance matrices of the fluorescence spectra for the two olive oil classes (EXTRA and LAMPANTE). 
Bright red areas correspond to regions of strong inter-wavelength covariances, notably around the main fluorescence peak and stray-light regions. 
Such covariance mismatches are sufficient, in high-dimensional space, to enable nearly perfect classification even when chemically meaningful information is absent.}
    \label{fig:cov1}
\end{figure*}

Figure \ref{fig:random_subsets} clearly shows how a flexible model (such as a random forest) is able to use statistical differences in the data to achieve a very good accuracy, even when it should not, from a chemical point of view, be able to.
This is naturally due to the fact that the different classes have different covariance matrices, as can be seen, for example, for EVOO and LOO in Figure \ref{fig:cov1}. Note that the large red areas in the lower right part of the covariances are related to the main peak, while the light red regions are due to stray light from the excitation LED. Note that for all the tests, we removed the Rayleigh peak from the spectra, but we wanted to include it into the covariances matrices for completeness.

\subsubsection{Feature Importance: Window Sweep}

\begin{figure*}[hbt]
    \centering
    \includegraphics[width=1\linewidth]{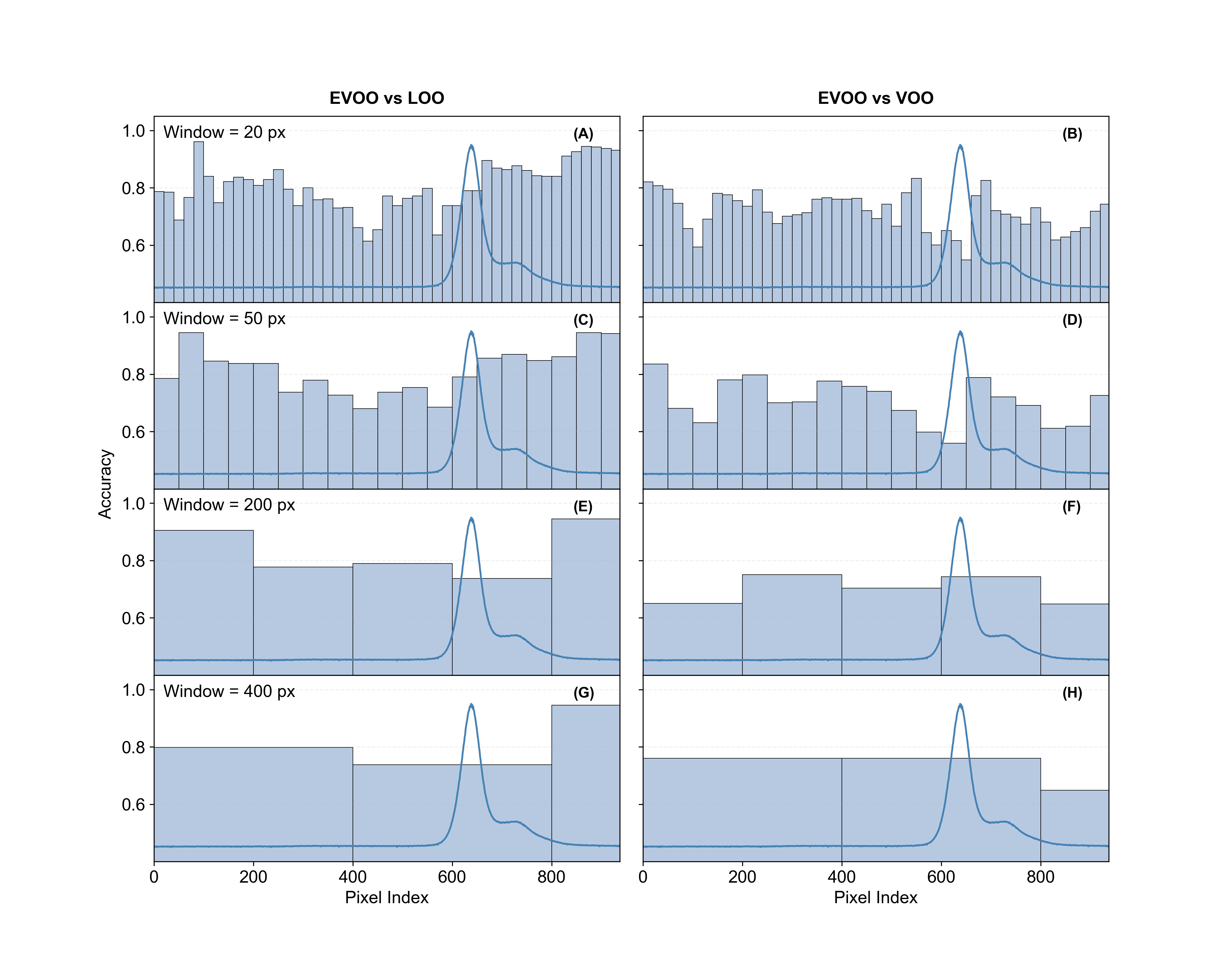}
    \caption{Results for experiment Ra4/Rb4. LOO-CV classification accuracy is mapped across the fluorescence spectrum (gray line) using non-overlapping windows of increasing size $W$. 
    Left Column (Panels A, C, E, G): Experiment EVOO vs. LOO. 
    Right Column (Panels B, D, F, H): Experiment EVOO vs. VOO. 
    The spectral region between 380--420 nm was explicitly removed to eliminate the Rayleigh scattering peak as a trivial discriminant. 
    Note that even at small window sizes (20--50 px), the models achieve $\sim$70--80\% accuracy in the chemically empty region (pixels 0--400). 
    As dimensionality increases to $W=400$ (Panels G and H), a stable accuracy plateau of $\sim$80\% emerges across the entire detector, confirming that high-dimensional noise distributions provide sufficient statistical information for class separation, independent of physical spectroscopic features.}
    \label{fig:sliding_window_comparison}
\end{figure*}

To further demonstrate how this phenomenon, we performed the experiment Ra4/Rb4 focussing on feature-importance selection. In spectroscopy, this is often done by identifying which spectral regions yield the highest classification accuracy. Since the optimal width of such regions is typically unknown, we considered several window sizes $W$, specifically 20, 50, 200, and 400 pixels.
For each width, we used only the spectral segment within the moving window as model input (resulting in input dimensions of 10, 50, 200, and 400 pixels, respectively) and systematically slid the window along the spectrum and evaluated the performance of a random forest classifier. The results can be seen in Figure \ref{fig:sliding_window_comparison} and can be summarised as follows.
\begin{itemize}
    \item \textbf{Universality Across Tasks:} Comparing experiment EVOO vs. LOO and experiment EVOO vs. VOO reveals a striking commonality. Despite the differing chemical complexities of these tasks, the classification accuracy in the chemically empty region ($\rho_1$, pixels 0--400) remains consistently high ($>80\%$) in both cases (Panels G and H). This suggests that the model is exploiting a universal instrumental covariance structure rather than task-specific chemical markers.
    
    \item \textbf{The Dimensionality Plateau:} The transition from $W=20$ to $W=400$ illustrates the impact of dimensionality on separability. In the $400$-pixel windows, the accuracy reaches a stable plateau that is indifferent to the underlying spectral profile. 
\end{itemize}

These results confirm that high importance in a ML model can often be an artefact of high-dimensional geometry. When a model can achieve $80\%$ accuracy using only randomised noise pixels (as shown in experiment Ra1/Rb1) or empty spectral windows, the standard interpretation of model weights as ``chemical signatures'' becomes invalid. We conclude that in 10$^3$-dimensional space, the most stable discriminant is frequently the global statistical fingerprint of the background, creating a deceptive ``path of least resistance'' that bypasses the intended chemical analysis.

\subsection{Feature Importance: SHAP}

A common approach in ML applied to spectroscopy is the use of additive feature attribution methods, such as SHAP (SHapley Additive Explanations) \cite{LundbergLee2017}, to validate model decisions. With the experiment Ra5/Rb5 we show that SHAP is subject to the same high-dimensional ``path of least resistance'' as the underlying classifier.
Mathematically, SHAP values are designed to distribute the total payout (the model's prediction) among the features (pixels). When classes are mutually singular due to infinitesimal shifts in the high-dimensional noise, the noise pixels collectively hold all the discriminatory power. Consequently, SHAP correctly identifies these pixels as the primary drivers of the success of the model. 

However, this identifies a statistical shortcut rather than a chemical signature. In high-dimensional spaces, shortcuts are more robust and easier for the model to minimise training loss than the complex, non-linear signals of chemical peaks. Thus, a high SHAP value in a noise-dominated region is not a sign of a hidden chemical feature; it is an empirical confirmation that the model has successfully exploited the geometric separability of the instrumental background.

\begin{figure*}[t!]
    \centering
    \includegraphics[width=1\linewidth]{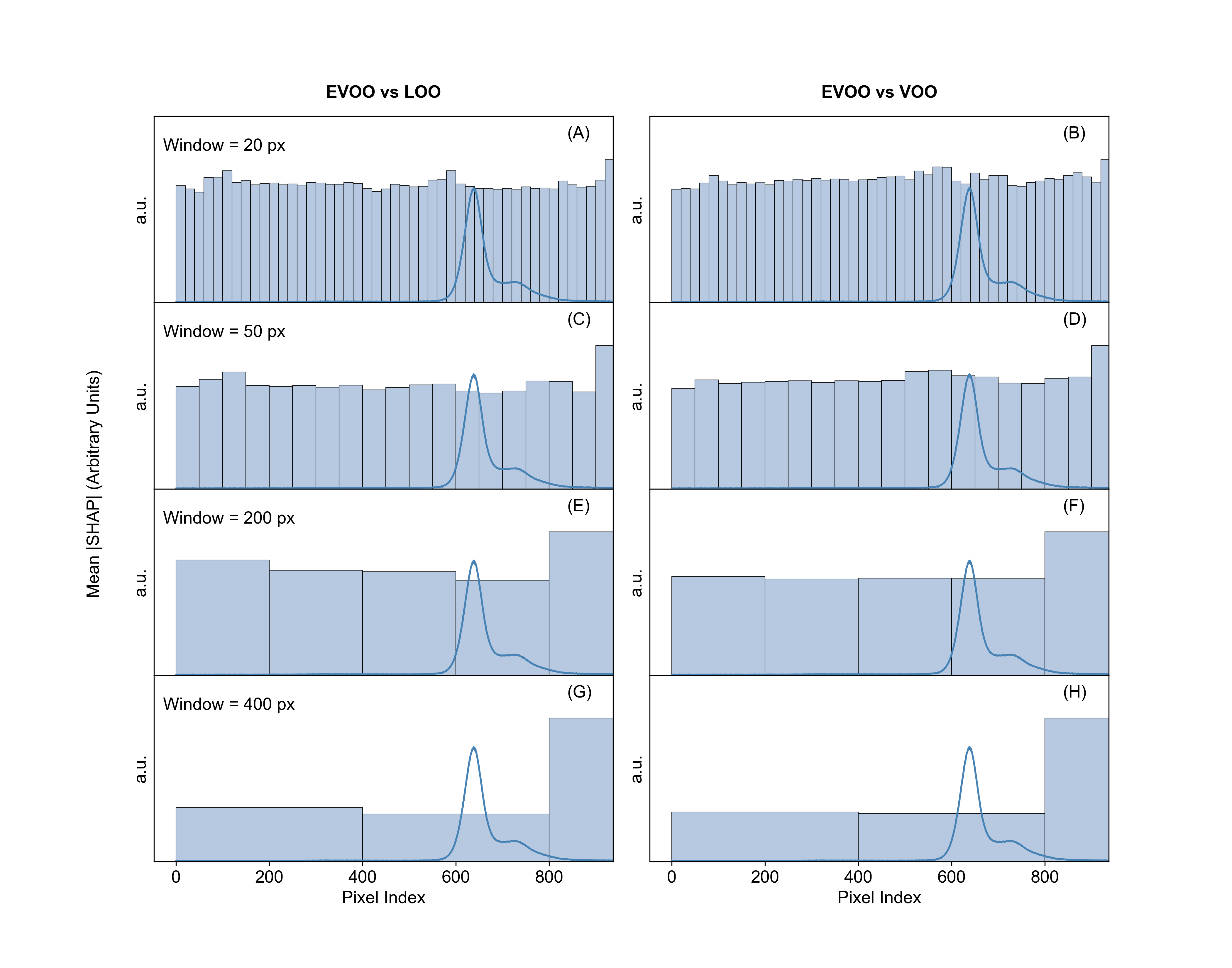}
    \caption{Results for experiment Ra5/Rb5. Regional Feature Attribution Map. Mean absolute SHAP values are presented in arbitrary units (a.u.) to facilitate the comparison of relative feature importance across varying window sizes ($W$).
    Left Column (Panels A, C, E, G): Experiment EVOO vs. LOO. 
    Right Column (Panels B, D, F, H): Experiment EVOO vs. VOO. 
    The spectral region between 380--420 nm was explicitly removed to eliminate the Rayleigh scattering peak as a trivial discriminant. The persistence of high attribution in the noise-only region (pixels 0–400) across both experimental tasks indicates a reliance on high-dimensional statistical shortcuts rather than physical chemical signatures.}
    \label{fig:shap}
\end{figure*}

To explain the model's decisions, we employed SHAP . For each window size $W \in \{20, 50, 200, 400\}$, we trained a random forest classifier on the localised spectral segment $X_{start:start+W}$.
 SHAP values were calculated using the \texttt{}{TreeExplainer} algorithm. To quantify the importance of a spectral window, we calculated the global mean absolute SHAP value. This metric allows us to map which spectral regions were used as primary discriminants by the model.
The results can be found in Figure \ref{fig:shap}.
The feature attribution analysis was performed on both both classification of EVOO vs. LOO and EVOO vs. VOO. As shown in Figure \ref{fig:shap}, the attribution profiles reveal a significant decoupling between the spectral regions identified as important from the model and the physical chemical signal.
Across all window sizes ($W=20$ to $W=400$ px), the model assigns high importance to spectral regions where the chemical signal is low or entirely absent. Notably, in the $400$-pixel window regime (Panels G and H), the importance assigned to the noise-only region (pixels 0--400) is comparable to, or even exceeds, the importance assigned to the primary fluorescence peaks (pixels 600--800). This indicates that the model is not relying on specific chemical markers, but is instead utilising the global high-dimensional background as a primary discriminant.

\section{Discussion and Practical Implications for Spectroscopic Modeling}
\label{sec:spectroscopy}

The results presented in this work demonstrate that the apparent success of machine learning in spectroscopy is often a consequence of the inherent high dimensionality of spectral data. In such spaces, even tiny distributional differences, such as small variations in noise or background, make the data perfectly separable as the number of spectral points grows. As a result, classifiers can achieve seemingly perfect accuracy without learning chemically meaningful features, relying instead on subtle artefacts or instrument–specific noise patterns. 

When feature selection or wavelength band selection is applied to spectroscopy data, the high dimensionality of spectra can produce misleading results. Since classifiers can exploit minute distributional differences, even in spectral regions that contain no physico-chemically meaningful information, commonly used importance approaches often highlight bands that are merely correlated with noise patterns or instrument artefacts. For instance, a random forest might consistently assign high importance to regions far from characteristic peaks, not because those wavelengths encode chemical signatures but because small statistical fluctuations in those regions suffice to separate classes in high-dimensional space. This effect can lead spectroscopists to misinterpret the outcome of ML models. A feature ranking that emphasises noise-driven regions can be taken as evidence of a new ``hidden'' marker, while in reality the model is simply exploiting spurious differences in baseline or detector noise. As a result, band-selection workflows risk reinforcing artefacts rather than guiding the discovery of meaningful chemical or physical features. This danger is especially acute when spectra are normalised or preprocessed, since those steps may amplify or redistribute noise in ways that make certain bands appear systematically discriminative.

Therefore, great caution is required when interpreting the output of band-importance methods. Any highlighted region should be cross-validated against established chemical knowledge or verified with independent measurements. Without this step, spectroscopists risk drawing incorrect conclusions, such as attributing predictive power to wavelength regions that carry no true spectroscopic signal. The findings of this study suggest that feature selection in spectroscopy, if performed without domain knowledge, can easily mislead and produce models that generalise poorly across instruments, conditions, or sample sets.

Dark signal and stray light must be mentioned in this context. In fact, they act as structured ``noise’’ that can differ by instrument, session, or acquisition order and can alone enable near‐perfect separation in high dimension and mislead band‐importance analyses (e.g., highlighting off‐peak regions with no chemical content). Models trained under such conditions may fail to generalise across instruments or setups, despite excellent internal validation. Practically, this calls for rigorous controls: randomise acquisitions across classes, replicate across instruments/sessions, evaluate with leave‐instrument/session‐out validation, and verify that accuracy collapses when noise statistics are equalised (e.g., per‐scan mean/variance standardisation or explicit dark/stray‐light correction). Only signals that remain discriminative under these checks should be interpreted as chemically meaningful.

Ultimately, this work should not be interpreted as a general refutation of machine learning in spectroscopy, but rather as a call for a more rigorous, evidence-based framework for model validation; we propose that high classification accuracy must be accompanied by regional sensitivity audits—such as the windowed SHAP analysis and global shuffle tests presented here—to ensure that model success is derived from verifiable chemical signatures rather than high-dimensional statistical shortcuts.


When applying machine learning to spectroscopy, it is essential to check whether models are separating classes based on chemically meaningful information or on trivial artefacts. A useful diagnostic is to test performance on wavelength regions that should be indistinguishable and contain no chemical signal; if the model still performs above chance, then separability is likely driven by noise or measurement artefacts.  

Preprocessing choices also play a critical role. Steps such as baseline subtraction or normalisation can unintentionally amplify or suppress noise patterns, creating the illusion of meaningful separation. Similarly, spectral band importance methods (e.g.\ feature maps from random forests, SVMs, or SHAP values) may highlight regions that correspond to noise rather than true peaks, and therefore these results should be interpreted with great caution.  

In conclusion, spectroscopists must remain aware that models trained on data from one instrument or measurement setup may not generalise to another. Retraining or re–validation is essential when changing experimental conditions. The safest approach is to combine machine learning with domain knowledge of peak positions, line shapes, and chemical constraints, and to begin with synthetic or well–characterised spectra where the discriminative features are known. This provides a baseline to ensure that models are learning physically relevant information rather than statistical quirks of the dataset.  

\subsection{Distinguishing Overfitting from High-Dimensional Separability}

It is essential to distinguish between classical overfitting and the phenomenon of high-dimensional separability described in this work. Although both can result in deceptively high accuracy, their underlying mechanisms differ.

\begin{itemize}
    \item \textbf{Overfitting} typically occurs when the complexity of a model (number of parameters) is too high relative to the number of samples $N$. In this state, the model "memorises" specific noise fluctuations in the training set that do not exist in the population.
    
    \item \textbf{High-Dimensional Separability} (the Feldman-Hájek effect) is a geometric property where two distributions become mutually singular as the number of dimensions $n$ increases. In this case, the model is not necessarily "memorising" noise; rather, it is correctly identifying that in $10^3$ dimensions, the classes occupy disjoint regions of space due to minute differences in their global covariance or mean.
\end{itemize}

A key diagnostic to distinguish the two is the \textit{rate of convergence} to perfect accuracy. In classical overfitting, accuracy usually improves as the number of samples $N$ decreases (making the ``memorisation'' easier). In contrast, high-dimensional separability is driven by the number of pixels $n$. As shown in our experiments (Figures 6 and 10), even with a fixed or increasing sample size, accuracy increases steadily as more spectral points are added. Furthermore, our ``shuffle'' experiments demonstrate that the model is exploiting global statistical distributions, which are properties of the population, rather than just local pixel-wise noise.

\subsection{Generality Across Spectroscopic Techniques}

Although real-world validation in this study uses fluorescence spectra, the theoretical framework grounded in the Feldman-Hájek theorem and the concentration of measure is inherently platform-independent. The phenomenon of high-dimensional separability is a property of the data's geometry rather than its physical origin. 
Consequently, these findings are highly relevant to other common techniques, such as Near-Infrared (NIR) or Raman spectroscopy. In NIR spectroscopy, specifically, spectral features are characterised by broad, overlapping features rather than sharp, distinct peaks. This low ``chemical contrast'' makes NIR models particularly susceptible to the ``path of least resistance'' described in this work: a high-dimensional classifier may easily bypass the subtle chemical signals in favor of infinitesimal but perfectly separable differences in the instrumental noise floor or baseline offsets.

\subsection{Model Complexity and Susceptibility to Dimensional Effects}

It is important to note that the theoretical framework presented here does not imply that all ML models are destined to fail in high-dimensional spectroscopic applications. Rather, it suggests that different architectures exhibit varying levels of susceptibility to these geometric effects. 
Highly flexible, non-linear models—such as deep neural networks or random forests, are particularly adept at identifying and exploiting subtle, high-dimensional covariance patterns. Because these models prioritise the minimisation of training error, they naturally gravitate toward the most statistically separable features, which, in high dimensions, are frequently instrumental artefacts or noise distributions rather than complex chemical signatures. In contrast, simpler linear models or those incorporating strong domain-specific regularisation may be less prone to ``meaningless'' success, provided they are constrained to look for physically plausible features.

Also it is important to note, that even in high dimensions and with clear statistical differences, it is possible that specific model classes will not reach a high accuracy. The fact that two classes are, in principle, perfectly separable it does not mean that every model can do that, or that it is an easy task. The deicision bounday may be too complex for specific model classess to detect, and thus even very flexible models might have a low accuracy in classification tasks, even if in high dimensions.

\section{Conclusions}
\label{sec:concl}

This study provides a solid theoretical and empirical refutation of the assumption that high classification accuracy is a sufficient proxy for a model learning from physico-chemical information in spectroscopy. By linking the \textit{Feldman-H\'{a}jek} theorem to experimental fluorescence data, we have shown that the ``Path of Least Resistance'' for a ML model to obtain high accuracy can be the background rather than the intended chemical signal. 

We hope that our findings serve as a useful framework for the field: we propose that the standard for ``model success'' must be elevated from simple cross-validation accuracy to a rigorous \textbf{Regional Sensitivity Audit}. The windowed SHAP importance maps, the global shuffle tests, and the physical feature-removal protocols developed here provide a possible blueprint for a new generation of ``physically-aware'' machine learning. By adopting these stress-tests, the spectroscopy community can safeguard against the publication of non-replicable ``phantom'' models and ensure that the power of artificial intelligence is harnessed to uncover genuine molecular insights rather than high-dimensional geometric artefacts.

Note that while this study utilises fluorescence spectra characterised by broad features and significant instrumental background, it is important to note that the impact of high-dimensional statistical shortcuts may vary in datasets containing more dense chemical information, such as the infrared fingerprint regions of biological tissues, where the higher 'chemical contrast' might offer more robust physical discriminants


\section*{Author contributions}
Umberto Michelucci: Conceptualisation, Data Curation, Methodology, Formal analysis, Investigation, Writing – original draft, Writing – review \& editing. Francesca Venturini: Conceptualisation, Methodology, Validation, Writing – original draft, Writing – review \& editing.

\section*{Conflicts of interest}
There are no conflicts to declare.

\section*{Data availability}

This study was carried out using synthetic data. The data generation is explained in the paper. The olive data were previously described \cite{venturini2021exploration} and are openly available in ``Dataset of Fluorescence Spectra and Chemical Parameters of Olive Oils'' at \url{https://data.mendeley.com/datasets/thkcz3h6n6/6}, DOI: 10.17632/thkcz3h6n6.6.


\section*{Acknowledgements}
 This project/research has received funding from the European Union’s Horizon Europe research and innovation programme under the Marie Skłodowska-Curie Actions (MSCA) Staff Exchanges, Grant Agreement No. 101236434 (SMARTOLIVE).




\bibliography{references} 
\bibliographystyle{unsrt} 
\end{document}